\documentclass[pdflatex,referee,sn-basic]{sn-jnl}

\usepackage{multirow}
\usepackage{geometry}  
\usepackage{booktabs}
\usepackage{siunitx}
\usepackage{threeparttable}
\usepackage{rotating}
\usepackage{color,xcolor}
\usepackage{gensymb}
\usepackage{subfig}

\usepackage{amssymb}
\usepackage{amsmath}
\usepackage{url}
\usepackage{algorithm}
\usepackage{algpseudocode}

\newcommand{\Rb}{\mathbb{R}}
\newcommand{\bfs}{\mathbf{s}}

\jyear{2023}%

\begin{document}

\title[Computationally efficient spatial regression]{A parsimonious, computationally efficient machine learning method for spatial regression}



\author*[1]{\fnm{Milan} \sur{\v{Z}ukovi\v{c}}}\email{milan.zukovic@upjs.sk}

\author[2]{\fnm{Dionissios T.} \sur{Hristopulos}}\email{dchristopoulos@tuc.gr}

\affil*[1]{\orgdiv{Department of Theoretical Physics and Astrophysics, Institute of Physics, Faculty of Science, \orgname{Pavol Jozef Šafárik University}, \orgaddress{\street{Park Angelinum 9}, \city{Ko\v{s}ice}}, \postcode{041 54},  \country{Slovak Republic}}}

\affil[2]{\orgdiv{Department of Electrical and Computer Engineering}, \orgname{Technical University of Crete}, \orgaddress{\street{Akrotiri Campus}, \city{Chania}, \postcode{73100}, \state{Crete}, \country{Greece}}}

\abstract{
We introduce the modified planar rotator method (MPRS), a physically inspired machine learning method for spatial/temporal regression.   MPRS is a non-parametric model which incorporates spatial or temporal correlations via short-range, distance-dependent  ``interactions'' without assuming a specific form for the underlying probability distribution. Predictions are obtained by means of a fully autonomous learning algorithm which employs equilibrium  conditional Monte Carlo simulations. MPRS is able to handle scattered data and arbitrary spatial dimensions. 
We report tests on various synthetic and real-word data in one, two and three dimensions which demonstrate that the MPRS prediction performance (without parameter tuning) is competitive with standard interpolation methods such as ordinary kriging and inverse distance weighting. In particular, MPRS is a particularly effective gap-filling method for rough and non-Gaussian data (e.g., daily precipitation time series). MPRS shows superior computational efficiency and scalability for large samples. Massive data sets involving millions of nodes can be processed in a few seconds on a standard personal computer.}


\maketitle



\section{Introduction}
\label{sec:intro}
The spatial prediction (interpolation) problem arises in various fields of science and engineering that study spatially distributed variables. In the case of scattered data, filling gaps  facilitates understanding of the spatial features, visualization of the observed process, and it is also necessary to obtain fully populated grids of spatially dependent parameters used in partial differential equations. Spatial prediction is highly relevant to many disciplines,
such as environmental mapping, risk assessment~\citep{christakos12} and environmental
health studies~\citep{christakos13},  subsurface hydrology~\citep{kitan97,rubin03}, mining~\citep{goov97}, and oil reserves estimation ~\citep{hohn00,hamzehpour06}. In addition, remote sensing images often include gaps with missing data (e.g., clouds, snow, heavy precipitation, ground vegetation coverage, etc.) that need to be filled~\citep{rossi94}. Spatial prediction is also useful in image analysis~\citep{winkler03,gui04} and signal processing~\citep{unser05,ramani06} including medical applications~\citep{parrot93,cao01}.

Spatial interpolation methods  in the literature include simple  deterministic approaches, such as inverse distance weighting~\citep{shepard1968two} and minimum curvature~\citep{sandwell87}, as well as the widely-used  family of  kriging estimators~\citep{cressie90}. The latter  are stochastic methods, with their popularity being due to  favorable statistical properties (optimality, linearity, and unbiasedness
under ideal conditions). Thus, kriging usually outperforms other interpolation methods in prediction accuracy. However, the computational complexity of kriging increases cubically with the sample size and thus becomes impractical or infeasible for large data sets. On the other hand, massive data are now ubiquitous due to modern sensing technologies such as radars, satellites, and lidar.

To improve computational efficiency,  traditional methods can be modified leading to some tolerable loss of prediction performance~\citep{cres08,furr06,ingram08,kauf08,marco18,zhong16}. With new developments in hardware architecture, another possibility is provided by parallel implementations using already rather affordable multi-core CPU and GPU hardware architectures~\citep{cheng13,guti14,hu15,pesq11}. A third option is to propose new prediction methods that are inherently computationally efficient.

One such approach employs Boltzmann-Gibbs random fields  to model spatial
correlations by means of short-range ``interactions'' instead of the empirical variogram (or covariance) function used in geostatistics~\citep{dth03,dthsel07,dth15}. This approach was later extended to non-Gaussian gridded data by using classical spin models~\citep{mz-dth09a,mz-dth09b,mz-dth13b,mz-dth18,zuko20gpu}. The latter were shown to be computationally efficient and competitive in terms of prediction performance with respect to several other interpolation methods. Moreover, their ability to operate without user intervention makes them ideal candidates for automated processing of large data sets on regular spatial grids, typical in remote sensing. Furthermore, the short-range (nearest-neighbor) interactions between
the variables allows parallelization and thus further increase in computational efficiency. For example, a GPU implementation of the modified planar rotator (MPR) model led to impressive speed-ups (up to almost 500 times on large grids), compared to single CPU calculations~\citep{zuko20gpu}.

The MPR method is limited to 2D grids, and its extension to scattered data is not  straightforward. In the present paper we propose the modified planar rotator for scattered data (MPRS). This new machine learning method can be used for scattered or gridded data in spaces with different  dimensions.  MPRS achieves even higher computational efficiency than MPR due to  full vectorization of the algorithm. This new approach does not rely a particular structure or dimension of the data location grid; it only needs the distances between each prediction point and a predefined number of samples in its neighborhood. This feature makes MPRS  applicable to scattered data in arbitrary dimensions. 

\section{The MPRS Model}
\label{sec:mprs}
The MPRS model exploits an idea initially used in the modified planar rotator (MPR) model~\citep{mz-dth18}. The latter was introduced for filling data gaps of continuously-valued variables distributed on 2D rectangular grids. The key idea is to map the data to continuous ``spins'' (i.e., variables defined in the interval $[-1,1]$), and then construct a model of spatial dependence by imposing interactions between  spins.  MPRS models such interactions even between scattered data and is thus applicable to both structured and unstructured (scattered) data over domains $D \subset \Rb^{d}$ where $d$ is any integer. 

\subsection{Model definition}

Let us consider  the random field $Z(\bfs;\omega) : \Rb^{d} \to \Rb$ which is defined over  a complete probability space $(\Omega,\mathcal{F},P)$. Assume that the data are sampled at points $G_{s}=\{ \bfs_{i} \in \Rb^{d} \}_{i=1}^{N} $ from the field $Z(\bfs;\omega)$.  The data set is denoted by $Z_{s}=\{z_{i} \in {\mathbb{R}} \}_{i=1}^{N}$, and the set of prediction points  by $G_{p}=\{\bfs_{p} \in \Rb^{d}\}_{p=1}^{P}$ so that $G_{s} \cup G_{p} = G$, $G_{s} \cap G_{p} = \emptyset$ (i.e., the sampling and prediction sets are disjoint), and $P+N = N_{G}$. The random field values at the prediction sites will be denoted by the set $Z_{p}$.

A Boltzmann–Gibbs probability density function (PDF) can be defined for the configuration $z(\bfs)$ sampled over  $G_s \subset \Rb^{d}$. The PDF  is governed by the Hamiltonian (energy functional) ${\mathcal H}(z_{G_s})$ and is given by the exponential form 
\begin{equation}
\label{Gibbs}
f_{Z}(z_{G_s}) ={\mathcal Z}^{-1} \exp \{-{\mathcal H}(z_{G_s})/k_BT \},
\end{equation}
where $z_{G_s} \triangleq\{z(\bfs): \, \bfs \in G_s \}$ is the set of data values at the sampling points, and ${\mathcal Z}$ is a normalizing constant (known as the partition function). In statistical physics, $T$ is the thermodynamic temperature and $k_B$ is the Boltzmann constant.  In the case of the MPRS model,  the product $k_BT$ represents a model parameter that controls the variance of the field.

A local-interaction Hamiltonian can in general be expressed as
\begin{align}
\label{Hamiltonian_MPRS}
{\mathcal H} = & - \langle J_{i,j} \,\Phi(z_{i},z_{j}) \rangle,
\end{align}
where $J_{i,j}$ is a location-dependent pair-coupling function and $\Phi(z_{i},z_{j})$ is a nonlinear function of the interacting values $z_{i}, z_{j}$. The notation $\langle \cdot \rangle$ implies a spatial averages defined by means of
\begin{align}
\label{average}
\langle A_{i,j} \rangle \triangleq & \sum_{i}\sum_{j \in \mathrm{neighb}(i)} A_{i,j}
\end{align} 
where $A_{i,j}$ is a two-point function,  and  $\mathrm{neighb}(i)$ denotes all $n_b$ sampling points in the interaction neighborhood of the $i$-th point. 

To define the local interactions in the MPRS model, the original data $z_i$ are mapped to continuously-valued ``spin'' variables represented by angles $\phi_i$  using the linear transformation
\begin{equation}
\label{map}
z_{i} \mapsto \phi_{i} = \frac{2\pi(z_{i}- z_{s,\min})}{z_{s,\max} - z_{s,\min}}\,,\; \textrm{for} \; z_{i} \in Z_{s},
\end{equation}

\noindent where $z_{s,\min}= \min_{i \in \{1,2, \ldots, N\}} Z_{s}$,  $z_{s,\max} = \max_{i \in \{1,2, \ldots, N\}} Z_{s}$,   and $\phi_{i} \in [0,2\pi]$, for $i=1, \ldots, N$.
The MPRS pairwise energy is given by the equation 
\begin{equation}
\label{eq:Phi-ij}
\Phi_{i,j}=\cos\phi_{i,j}, \; \textrm{where} \; \phi_{i,j}=(\phi_{i}-\phi_{j})/2\,.
\end{equation}





In order to fully determine interactions between scattered data, the coupling function $J_{i,j}$ needs to be defined.  It is reasonable to assume that the strength of the interactions diminishes with increasing distance.  Hence, we adopt an exponential decay of the interactions between two points $i$  and $j$, i.e., 
\begin{equation}
 \label{eq:exp_dec}
 J_{i,j} =J_0\exp(-r_{i,j}/b_i).
\end{equation}
In the coupling function~\eqref{eq:exp_dec},  the constant $J_0$ defines the maximum intensity of the interactions,  $r_{i,j}=\lVert \bfs_{i} - \bfs_{j} \rVert$ is the pair distance, and the locally adaptive bandwidth parameter $b_i$ is specific to each prediction point and reflects the sampling configuration in the  neighborhood of the point $\bfs_i$. Note that although the coupling function decays smoothly, the energy~\eqref{Hamiltonian_MPRS} embodies interactions only between points that are inside the specified neighborhood of each point $\bfs_i$. 

\begin{figure}[t!]
\centering
\subfloat{\includegraphics[scale=0.5,clip]{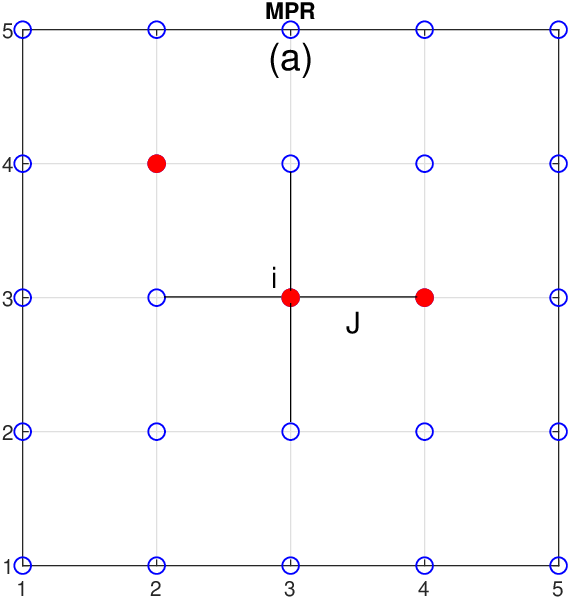}\label{fig:schem_mpr}}\hspace{15mm}
\subfloat{\includegraphics[scale=0.5,clip]{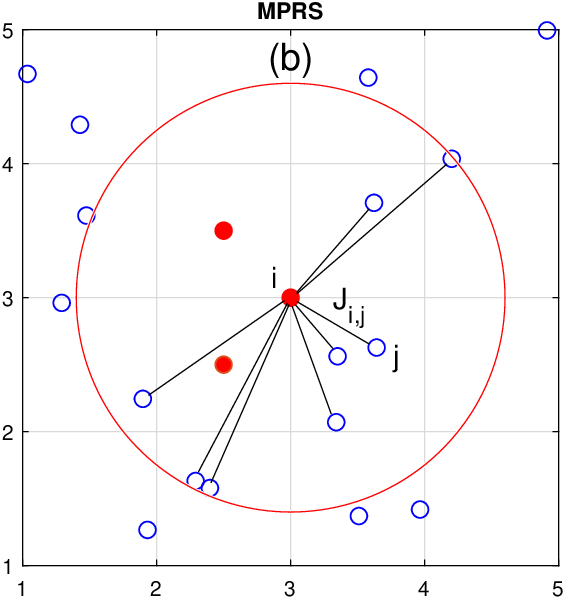}\label{fig:schem_mprs}}
\caption{Schematic illustration of the interactions of $i$th prediction point with (a) its four nearest neighbors (including sampling and prediction points) via the constant interaction parameter $J$ in MPR and (b) its $n_b=8$ nearest neighbor (only sampling) points via the mutual distance-dependent interaction parameter $J_{i,j}$ in MPRS. Blue open and red filled circles denote sampling and prediction points, respectively, and the solid lines represent the bonds.}
  \label{fig:schem}
\end{figure}

The interactions in the MPRS model are schematically illustrated and compared with the MPR interactions in Fig.~\ref{fig:schem}. The diagram clarifies how the MPRS method extends the coupling to scattered data sets. 

In MPRS, regression is accomplished by means of a conditional simulation approach which is described below.  To predict the value of the field at the points in $G_p$, the energy function~\eqref{Hamiltonian_MPRS} is extended to include the prediction points, i.e., we use ${\mathcal H}(z_{G})= {\mathcal H}(z_{G_s} \cup\, z_{G_p})$.  In ${\mathcal H}(z_{G})$ we restrict interactions between each prediction point and its sample neighbors (i.e., we neglect interactions between prediction points) in order to allow vectorization of the algorithm which enhances computational performance.  In practice, omitting prediction-point interactions does not impact significantly the prediction. Then, the Hamiltonian comprises two parts: one that involves only sample-to-sample interactions and one that involves interactions of the  prediction points with the samples in their respective neighborhood. Since the sample values are fixed, the first part contributes an additive constant, while the important (for predictive purposes) contribution comes from the second part of the  the energy.  The latter represents a summation of the contributions from all $P$ prediction points.

The optimal values of the spin angles $\phi_p$ at $\bfs_{p} \in G_{p}$ can then be determined by finding the configurations which maximize the Boltzmann-Gibbs PDF~\eqref{Gibbs}, where the energy is now replaced with ${\mathcal H}(z_{G})$. If $T=0$, the PDF is maximized by the configuration which minimizes the total energy ${\mathcal H}(z_{G})$, 
i.e., 
\begin{equation}
\label{eq:phip-predict-0}
\hat{\phi}_{p} = \arg\min_{\phi_p} {\mathcal H}(z_{G}), \; p=1, \ldots, P\,.
\end{equation}

However, for $T \neq 0$, there can exist many configurations $\{ \phi_{p}\}_{p=1}^{P}$ that lead to the same energy ${\mathcal H}(z_{G})=E$.  Assuming that $\Omega(E)$ is the total number of configurations with energy $E$, the probability $P(E)$ of observing $E$ is $P(E) \propto \Omega(E)\, \exp\left[-{\mathcal H}(z_{G})/k_B T\right]$.  Equivalently, we can write this as follows
\begin{equation}
\label{ProbE}
P(E) \propto {\mathrm{e}}^{- \frac{1}{k_B T} \left[{\mathcal H}(z_{G}) - k_B T \log\Omega(E)\right]}\,.
\end{equation}
Taking into account that $S(E)= k_{B}\log\Omega(E)$ is the \emph{entropy} that corresponds to the energy $E$, the exponent of~\eqref{ProbE} becomes proportional to 
the \emph{free energy}: $F(E)= {\mathcal H}(z_{G}) - T \, S(E)$.  
Thus, for $T \neq 0$ an ``optimal configuration'' is obtained by means of
\begin{equation}
\label{eq:phip-predict}
\hat{\phi}_{p} = \arg\min_{\phi_p} \left[ {\mathcal H}(z_{G}) - T \, S(E) \right]\,, \quad p=1, \ldots, P\,.
\end{equation}
The minimum free energy corresponds to the thermal equilibrium state. In practice, the latter can be achieved in the long-time limit by constructing a sequence (Markov chain) of states using one of the legitimate updating rules, such as the Metropolis algorithm~\citep{Metropolis53}, as shown in Section~\ref{ssec:learning-prediction}. 
 

Finally, the MPRS prediction at the sites $\bfs_p \in G_{p}$ is formulated by inverting the linear transform~\eqref{map}, i.e.,  
\begin{equation}
\label{eq:invert-predict}
\hat{z}_{p}  = \left( z_{s,\max} - z_{s,\min}\right) \frac{\hat{\phi}_{p}}{2\pi}  + z_{s,\min}\,,\; \textrm{for} \; p =1, \ldots, P \,.
\end{equation}

The MPRS model is fully defined in terms of the equations~\eqref{Gibbs}-\eqref{eq:invert-predict}.

\subsection{Setting the MPRS Model Parameters and Hyperparameters}
\label{subsec:mprs_sim}
The MPRS  learning process involves the model parameters and a number of hyperparameters which control the approach of the model to an equilibrium probability distribution. The model parameters include the number of interacting neighbors per point, $n_b$,  the decay rate vector $\mathbf{b}=(b_{1}, \ldots, b_{P})^\top$  used in the exponential  coupling function~\eqref{eq:exp_dec}, the prefactor $J_0$, and the simulation temperature $T$; the ratio of the latter two  sets the interaction scale via the reduced coupling parameter $J_0/k_BT$.  Thus, in the following we refer to the ``simulation temperature'' ($T$) as  shorthand for the dimensionless ratio $k_BT/J_0$.  In addition, henceforward energy functions ${\mathcal H}$ are calculated with $J_0=1$.

In order to optimize computational performance, after experimentation with various data sets
we   set the model parameters to the default values $n_b=8$ (for all prediction points) and $T=10^{-3}$; the decay rates $\{ b_p \}_{p=1}^{P}$ are estimated as the median distance between the $p$-th prediction point and its four nearest sample neighbors. These choices are  supported by (i) the expectation of increased spatial continuity for low $T$ and (ii) experience with  the MPR method. In particular, MPR tends to perform better at very low $T$ (i.e., for $T \approx 10^{-3}$). In addition, using  higher-order  neighbor interactions ($n_{b}=8$ neighbors,  i.e., nearest- and second-nearest neighbors on the square grid)  improves the smoothness of the regression surface. The definition of the  decay rate vector  $\mathbf{b}$ enables it to adapt to potentially uneven spatial distribution of samples around prediction points.

Our exploratory tests showed that the prediction performance is not sensitive to the default values defined above. For example,  setting $n_b =4$ or  increasing (decreasing)  $T$ by one order of magnitude, we obtained very similar results as for the default parameter choices. Nevertheless, we tested the default settings on various data sets and verified that even if they are not optimal, they still provide competitive performance. 

The MPRS hyperparameters are used to control the learning process. The static hyperparameters are listed in Section 1.3.1 of Algorithm~\ref{algo:mpr-simul}. Below, we discuss their definition, impact on prediction performance, and setting of default values. The number of equilibrium configurations, $M$, is arbitrarily set to $100$. Smaller (larger) values  would  increase (decrease) computational performance and decrease (increase)  prediction accuracy and precision. The frequency of equilibrium state verification is controlled by  $n_f$ which is set to $5$. Lower $n_f$ increases the frequency and thus slightly decreases the simulation speed but it can lead to earlier detection of the equilibrium state.  
In order to test for equilibrium conditions, we need to check the slope of the energy evolution curve: in the equilibrium regime the curve is flat, while it has an overall negative slope in the relaxation (non-equilibrium) regime. However, the fluctuations present in equilibrium at $T \neq 0$ imply that the calculated slope will always quiver around zero.  To compensate for the fluctuations, we fit the energy evolution curve with a Savitzky-Golay polynomial filter of degree equal to one using a window that contains $n_{\textrm{fit}}=20$ points.  This produces a smoothed curve and a more robust estimate of the slope. Larger values of $n_{\textrm{fit}}$ are likely to cause undesired mixing of the relaxation and the equilibrium regimes.

The maximum number of Monte Carlo sweeps, $i_{\max}$, is optional and can be set to prevent very long equilibration times, lest the convergence is very slow. Due to the  efficient hybrid algorithm employed its practical impact is minimal. The target acceptance ratio of Metropolis update, $A_{\textrm{targ}}$, and the variation rate of perturbation control factor, $k_a$, are set to $A_{\textrm{targ}}=0.3$ and $k_a=3$. Their role is to prevent the Metropolis acceptance rate (particularly at low $T$) to drop to very low values, which would lead to computational inefficiency. Finally, the simulation  starts from some initially selected state of the spin angle configuration. Our tests showed that different choices, such as uniform (``ferromagnetic'') or random (``paramagnetic'') initialization produced similar results. Therefore, we use  as default the random state comprising  spin angles drawn from the uniform distribution in $[0,2\pi]$. While it is in principle possible to adjust the hyperparameters for optimal prediction performance, using default values enables the autonomous operation of the algorithm and controls the computational efficiency. The dynamic hyperparameters, listed in Section 1.3.2 of Algorithm~\ref{algo:mpr-simul}, increase the flexibility of the algorithm by automatically adapting  to the current stage of the simulation process.

\begin{algorithm}[t!]
\begin{small}
\caption{MPRS learning procedure. The algorithm involves the  \textsc{Update} function which is described in Algorithm~\ref{algo:mpr-relax}. $\mathbf{\Phi}_{s}$ is the vector of known spin values at the sample sites. $\mathbf{\hat{\Phi}}$  represents the vector of estimated spin values at the prediction sites. $G(\cdot)$ is the transformation from the original field to the spin field and $G^{-1}(\cdot)$ is its inverse. $\mathbf{\hat{Z}}(j)$, $j=1, \ldots, M$ is the $j$-th realization of the original field. $\mathbf{U}(0,2\pi)$ denotes a vector of random numbers from the uniform probability distribution in $[0, 2\pi]$. SG stands for Savitzky-Golay filter.}
\label{algo:mpr-simul}
\begin{algorithmic}
\State 1: \emph{Setting of parameters and hyperparameters}
\State 1.2: Model and simulation: $n_{b}$: $\#$ of bonds between each point and neighboring samples; $T$: simulation temperature; $\mathbf{b}$: vector of decay rates;
\State 1.3: Setting of static hyperparameters 
\State 1.3.1: Set the following to fixed values: $M$: $\#$ of equilibrium configurations (length of averaging sequence) for statistics collection; $i_{\max}$: maximum $\#$ of MC sweeps in non-equilibrium stage; $A_{\textrm{targ}}$: target acceptance ratio of Metropolis update; $k_a$: variation rate of perturbation control factor $a$; $n_{f}$: verification frequency of equilibrium conditions; $n_{\textrm{fit}}$: $\#$ of fitting points of energy evolution function;
\State 1.3.2: Initialize dynamic hyperparameters: slope of energy evolution function $ k(0) \gets - 1$; spin perturbation control factor $ a(0) \gets 1$; simulated state counter $ i \gets 0$;

\State 2: \emph{Data transformation}
\State  2.1: $\mathbf{\Phi}_{s} \gets G(\mathbf{Z}_{s})$  using~\eqref{map} \Comment Set data spin angles

\State 3: \emph{Interaction neighborhood}
\State 3.1: Identify $n_b$ neighbors \Comment For each prediction point find $n_b$ neighbors from samples and calculate distances

\State 4: \emph{Non-equilibrium spin relaxation procedure}
\State 4.1:  $\mathbf{\hat{\Phi}}(0) \gets \mathbf{U}(0,2\pi)$  \Comment Initialize spin values at prediction points 
\While{$[k(i)<0] \wedge [i \le i_{\max}]$ }    \Comment Spin updating with Metropolis sweep
\State 4.2: \Call{Update}{$\mathbf{\hat{\Phi}}(i+1), \mathbf{\hat{\Phi}}(i), a(i),T$}
\If{$A < A_{\textrm{targ}}$} \Comment Check Metropolis acceptance ratio
\State 4.3: $a(i+1) \gets 1+(i+1)/k_a$ \Comment Update perturbation control factor
\EndIf
\State 4.4: Calculate $E(i+1) \gets \mathcal{H}(G)$ \Comment Obtain current energy
\If{$[i \geq n_{\textrm{fit}}] \wedge [i \equiv 0 \pmod{n_f}]$} \Comment Check frequency of slope update of $E$
\State 4.5:  $k(i+1) \gets \mathrm{SG}$ \Comment Update slope of $E$ using last $n_{\textrm{fit}}$ values
\EndIf
\State 4.6: $ i \gets i +1$ \Comment Update MC counter
\EndWhile

\State 5. \emph{Equilibrium state simulation}
\State 5.1: $\mathbf{\hat{\Phi}}^{\mathrm{eq}}(0) \gets \mathbf{\hat{\Phi}}(i)$ \Comment Initialize the equilibrium state
\For{$j=0, \ldots, M-1$}
\State 5.2: \Call{Update}{$\mathbf{\hat{\Phi}}^{\mathrm{eq}}(j+1), \mathbf{\hat{\Phi}}^{\mathrm{eq}}(j), 1,T$} \Comment Generate equilibrium realizations
\State 5.3: $\mathbf{\hat{Z}}(j+1)  \gets G^{-1}\left[\mathbf{\hat{\Phi}}^{\mathrm{eq}}(j+1)\right]$ \Comment Back-transform spin states
\EndFor
\State 6: \Return Statistics of $M$ realizations $\mathbf{\hat{Z}}(j), \, j=1, \ldots, M$

\end{algorithmic}
\end{small}
\end{algorithm}

\begin{algorithm}[t!]
\begin{small}
\caption{Restricted Metropolis updating algorithm (non-vectorized version). $\mathbf{\hat{\Phi}}^{\mathrm{old}}$ is the
initial spin state, and $\mathbf{\hat{\Phi}}^{\mathrm{new}}$ is the new spin state. $\mathbf{\hat{\Phi}}^{\mathrm{old}}_{-p}$ is the initial spin
state excluding the point labeled by $p$. $U(0,1)$ denotes  the uniform probability distribution in $[0, 1]$.}
\label{algo:mpr-relax}
\begin{algorithmic}
\Procedure{Update}{$\mathbf{\hat{\Phi}}^{\mathrm{new}},\mathbf{\hat{\Phi}}^{\mathrm{old}},a,T$}
\For{$p=1, \ldots, P$}  \Comment Loop over prediction sites
\State 1:   $u \gets U(0,1)$ \Comment Generate uniform random number
\State 2:   ${\hat{\Phi}''}_p \gets {\hat{\Phi}'}_p + 2\pi (u -0.5)/a \pmod{2\pi}$ \Comment Propose spin update
\State 3:   $ \Delta \mathcal{H} = \mathcal{H}({\hat{\Phi}''}_p, \mathbf{\hat{\Phi}}^{\mathrm{old}}_{-p}) - \mathcal{H}({\hat{\Phi}'}_p, \mathbf{\hat{\Phi}}^{\mathrm{old}}_{-p})$ \Comment Calculate energy change
\State 4:  $AP = \min\{1,\exp(-\Delta \mathcal{H}/T)\}$ \Comment Calculate acceptance probability
\State 5:  $\mathbf{\hat{\Phi}}^{\mathrm{new}}_{-p} \gets \mathbf{\hat{\Phi}}^{\mathrm{old}}_{-p}$ \Comment Perform Metropolis update
            \If{$AP > r \gets U(0,1)$}
            \State 5.1: {$\hat{\Phi}_p^{\mathrm{new}} \gets {\hat{\Phi}''}_p$}    \Comment Update the state
            \Else
            \State 5.2: {$\hat{\Phi}_p^{\mathrm{new}} \gets {\hat{\Phi}'}_p$} \Comment Keep the current state
\EndIf
\EndFor \Comment End of prediction loop
\State 6: \Return $\mathbf{\hat{\Phi}}^{\mathrm{new}}$  \Comment Return the ``updated'' state
\EndProcedure

\end{algorithmic}
\end{small}
\end{algorithm}

\subsection{Learning ``Data Gaps'' by Means of Restricted Metropolis Monte Carlo} 
\label{ssec:learning-prediction}

MPRS predictions of the values $\{ \hat{z}_p \}_{p=1}^{P}$ are based on conditional Monte Carlo simulation. Starting with initial guesses for the unknown values, the algorithm updates them continuously aiming to approach an equilibrium state which minimizes the free energy (see Eq.~\ref{eq:phip-predict}). The key to the computational efficiency of the MPRS algorithm is fast relaxation to  equilibrium. This is achieved using the restricted Metropolis algorithm, which is particularly efficient at very low temperatures, such as the presently considered $T \approx 10^{-3}$, where the standard Metropolis updating is inefficient~\citep{loison2004canonical}. 

The classical Metropolis algorithm~\citep{Metropolis53,Robert99,dth20} proposes random changes of the spin angles at the prediction sites (starting from an arbitrary initial state).  The proposals are accepted if they lower the energy ${\mathcal H}(z_{G; \rm{curr}})$, while otherwise they are accepted with probability $p=\exp[-{\mathcal H}(z_{G; \rm{prop}})/T + {\mathcal H}(z_{G; \rm{curr}})/T]$, where $z_{G; \rm{curr}}$ is the current and $z_{G; \rm{prop}}$ the proposed states. 
The restricted Metropolis scheme generates a proposal spin-angle state according to  $\phi_{\rm{prop}}=\phi_{\rm{curr}}+\alpha(r-0.5)$, where $r$ is a uniformly distributed random number $r \in [0,1)$ and $\alpha=2\pi/a \in (0,2\pi)$. The hyperparameter $a \in [1, \infty)$ controls the \emph{spin-angle perturbations}. The value of $a$ is  dynamically tuned during the equilibration process to maintain the acceptance rate close to the target set by the \emph{acceptance rate hyperparameter} $A_{\rm{targ}}$. Values of $a \approx 1$ allow bigger perturbations of the current state, while  $a \gg 1$ leads to proposals closer to the current state.  

To achieve vectorization of the algorithm and high computational efficiency, we assume that interactions occur between prediction and sampling points in the vicinity of the former but not among prediction points. Moreover, perturbations  can be performed simultaneously by means of a single sweep for all the prediction points, which increases  computational efficiency (e.g., in the case of the MPR method two sweeps are required).

The learning procedure begins at an initial state ascribed to the prediction points, while the sampling points retain their values throughout the simulation.\footnote{If a prediction point coincides with a sample location, the MPRS algorithm allows the user to choose whether the sample value will be respected or  updated. Thus, in the former (latter) case MPRS is an exact (inexact) interpolator.} The prediction points can be initially assigned random values drawn from the uniform distribution. It is also possible to assign values based on neighborhood relations, e.g., by means of nearest neighbor interpolation. 

\begin{figure}[t!]
\centering
\includegraphics[scale=0.5,clip]{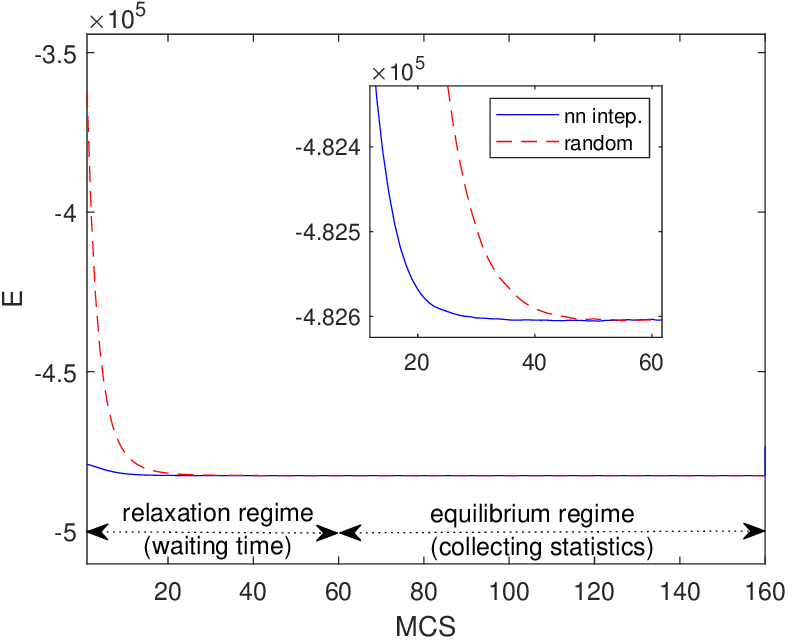}
\caption{Energy evolution curves starting from random (red dashed curve) and nearest-neighbor interpolation (blue solid curve) states. The simulations are performed on Gaussian synthetic data with $m= 150$, $\sigma=25$ and Whittle-Mat\'{e}rn covariance model WM($\kappa=0.2,\nu=0.5$), sampled at $346,030$ and predicted at  $702,546$ scattered points (non-coinciding with the sampling points) inside a square domain of length $L=1,024$. The inset shows a detailed view focusing on the nonequilibrium (relaxation) regime.}
\label{fig:ene_evol}
\end{figure}

Our tests showed that the initialization has marginal impact on  prediction performance but opting for the latter option tends to shorten the relaxation process and thus increases computational efficiency.  In Fig.~\ref{fig:ene_evol} we illustrate the evolution of the energy (Hamiltonian) ${\mathcal H}(z_{G})$  towards equilibrium using random   and nearest-neighbor initial states.  The curves represent interpolation on Gaussian synthetic data with   Whittle-Mat\'{e}rn covariance  (as described in  Section~\ref{subsec:results_synth}). The initial energy under random initial conditions differs significantly from the equilibrium value; thus the relaxation time (measured in MC sweeps), during which the energy exhibits a decreasing trend, is somewhat longer ($\approx$60 MCS) than for the nearest-neighbor initial conditions ($\approx$40 MCS). Nevertheless, the curves eventually merge and level off at the same equilibrium value.  
In order to automatically detect the crossover to equilibrium, i.e. the flat regime of the energy curve, the energy is periodically tested every $n_f$ MC sweeps, and the variable-degree polynomial Savitzky-Golay (SG) filter is applied~\citep{savgol64}. In particular, after each $n_f$ MC sweeps the last $n_{\textrm{fit}}$ points of the energy curve are fitted to test whether the slope (decreasing trend) has disappeared.

The MPRS predictions on $G_p$ sites are based on mean values obtained from $M$ states that are generated  via restricted Metropolis updating in the equilibrium regime. The hyperparameter $M$ thus controls the \emph{length of the averaging sequence.} The default value used herein is $M=100$.  Alternatively, the $M$ values  can be used to derive the \emph{predictive distribution} at each point on $G_p$. The entire MPRS prediction method is summarized in Algorithms~\ref{algo:mpr-simul} and~\ref{algo:mpr-relax}.

\section{Study Design  for Validation of MPRS Learning Method}
\label{sec:valid}

The prediction performance of the MPRS learning algorithm is tested on various 1D, 2D, and 3D data sets.  In 2D we use synthetic and real spatial data (gamma dose rates in Germany, heavy metal topsoil concentrations in the Swiss Jura mountains, Walker lake pollution, and atmospheric latent heat data over the Pacific ocean).  For 1D data we use time series of temperature and precipitation.  Finally, in 3D we use soil data.  The MPRS performance in 1D and 2D is compared with  ordinary kriging (OK) which under suitable conditions is an optimal spatial linear predictor~\citep{kitan97,cressie90,wack03}, and in 3D with inverse distance weighting (IDW). 

We compare prediction performance using different validation measures (see Table~\ref{tab:val-measures}). The complete data sets are randomly split into disjoint training and validation subsets. In most cases, we generate $V=100$ different training-validation splits.   Let $z(\bfs_p)$ denote the true value and $\hat{z}^{(v)}(\bfs_p)$ its estimate at $\bfs_p$ for the configuration $v=1, \ldots, V$. The \emph{prediction error} $\epsilon^{(v)}(\bfs_p)= z(\bfs_p) - \hat{z}^{(v)}(\bfs_p)$ is used to define validation measures over all the training-validation splits as described in Table~\ref{tab:val-measures}.

\begin{table}[!ht]
\caption{Validation measures used to assess the prediction performance of MPRS and other methods. The measures are defined as double averages: the first average is over the sites of the validation set while the second average is over all $V$ training-validation splits. The $\overline{z_{p}}$ and $\sigma_{z;p}$ are the mean and standard deviation of the validation values; $\langle\hat{z}^{(v)}_{p}\rangle$ and $\sigma^{(v)}_{\hat{z};p}$ are, respectively, the mean and standard deviation of the predictions for the $v$-th split. If $V=1$  the initial ``M'' in the validation measures can be dropped.}
\label{tab:val-measures}\centering
\renewcommand{\arraystretch}{2}
 \begin{tabular}{c|c}
    \toprule
     Validation measure    &  Definition \\
     \midrule
       Mean absolute error  & ${\mathrm{MAE}} = \frac{1}{V}\sum_{v=1}^{V}\frac{1}{P}\sum_{\bfs_{p} \in G_{p}} \lvert \,\epsilon^{(v)}(\bfs_{p}) \,\rvert$
        \\
       \hline
       Mean absolute relative error & ${\mathrm{MARE}} =\frac{1}{V}\sum_{v=1}^{V}\frac{1}{P} \sum_{\bfs_{p} \in G_{p}} \frac{\lvert \,\epsilon^{(v)}(\bfs_{p}) \rvert} {z(\bfs_{p})}$
       \\[1ex]
       \hline
       Mean root square error & ${\rm MRSE} =\frac{1}{V}\sum_{v=1}^{V}\sqrt{\frac{1}{P}\sum_{\bfs_{p} \in G_{p}}\,\left[ \epsilon^{(v)}(\bfs_{p})\right]^2}$
        \\[1ex]
       \hline
       Mean Pearson correlation coefficient & $MR=\frac{1}{V}\sum_{v=1}^{V}\frac{\sum_{p=1}^{P} \left(z(\bfs_p) - \overline{z_{p}} \right) \left( \hat{z}^{(v)}(\bfs_p) - \langle\hat{z}^{(v)}_{p}\rangle\right)}{\sigma_{z;p}\, \sigma^{(v)}_{\hat{z};p}}$
       \\
       \bottomrule
    \end{tabular}
\end{table}


To assess the computational efficiency of the methods tested we record the   CPU time, $t_{\mathrm{cpu}}$ for each split.  The mean computation time $\langle t_{\mathrm{cpu}} \rangle$ over all training-validation splits is then calculated. 
The MPRS interpolation method is implemented in  Matlab\circledR\ R2018a running on a desktop computer with 32.0 GB RAM and  Intel\textregistered Core\texttrademark2 i9-11900 CPU processor with a 3.50 GHz clock. 
Both OK and IDW  are applied at each prediction point using the entire training set (without defining a search neighborhood).

\section{Results}
\label{sec:results}

\subsection{Synthetic 2D data}
\label{subsec:results_synth}

Synthetic data are generated  from Gaussian, i.e., $Z \propto N(m = 150, \sigma = 25)$, and lognormal, i.e.,  $\ln Z \propto N(m = 0, \sigma)$ spatial random fields (SRF) using the spectral method for irregular grids~\citep{pardo1993fourier}. The spatial dependence is imposed by means of the  Whittle-Mat\'{e}rn (WM) covariance given by
\begin{equation}
\label{mate} C_{Z} (\|\mathbf{h}\|)=
\frac{{2}^{1-\nu}\, \sigma^{2}}{\Gamma(\nu)}(\kappa \,
\|\mathbf{h}\|)^{\nu}K_{\nu}(\kappa \, \|\mathbf{h}\|),
\end{equation}
where $\|\mathbf{h}\|$ is the Euclidean two-point distance, $\sigma^2$ is the variance, $\nu$ is the smoothness parameter, $\kappa$ is the
inverse autocorrelation length, and $K_{\nu}(\cdot)$ is the modified Bessel function of the second kind of order $\nu$. Hereafter,  we  use the abbreviation WM($\kappa,\nu)$ for such data. 
We focus on data with $\nu \leq 0.5$, which is appropriate for modeling rough spatial processes such as soil data~\citep{Minasny05}.

Data are generated at $N=10^3$ random locations within a square domain of size $L \times L$, where $L=50$. Assuming $tr$ represents the percentage of training points, from each realization we remove $\lfloor tr N \rfloor$ points to use as the training set. The predictions are cross validated with the actual values at the remaining locations. For various $tr$ values we generate $V=100$ different sampling configurations. 

The cross-validation measures obtained by  MPRS and OK  for $tr=0.10$ are summarized in Table~\ref{table:synth}. OK produces smaller errors and larger $MR$ than MPRS. However, the relative differences are typically $\lesssim 3\%$. On the other hand, the CPU time of MPRS is about 8 times smaller than for OK. 

\begin{table}[t!]
\caption{Cross-validation measures for MPRS and OK  based on 100 realizations of a Gaussian random field with mean $m= 150$,  standard deviation $\sigma=25$ and covariance model WM($\kappa=0.2,\nu=0.5$), sampled at $1,000$ scattered points inside a square of length $L=50$. 100 points are randomly selected as the training set ($tr=0.10$) and the values at the remaining 900 points are predicted.} 
\label{table:synth}
\centering
\begin{tabular}{lccccc}
\hline\noalign{\smallskip}
\hfil Method & \hfil MAE & \hfil MARE (\%) & \hfil RMSE & \hfil $MR$ (\%) & \hfil $\langle t_{\mathrm{cpu}} \rangle$ (s)  \\ \hline
  MPRS & $14.70$ &  $ 10.21 $&   $ 18.67 $&   $ 62.74$&    $ 0.02 $\\
  OK & $14.33$ &   $ 9.93 $&   $ 18.27 $&   $ 64.74$&    $ 0.16 $\\
\hline
\end{tabular}
\end{table}

\begin{figure}[t!]
\centering
\subfloat{\includegraphics[scale=0.48,clip]{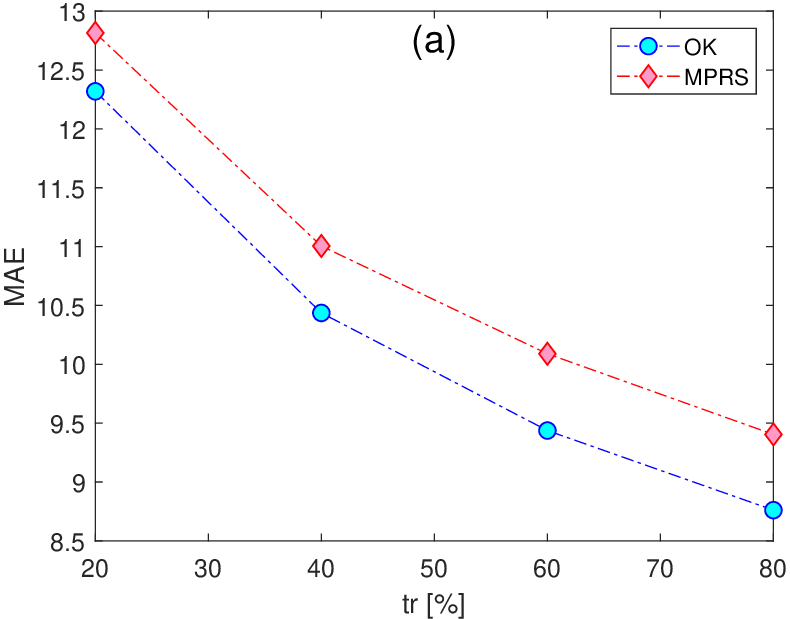}\label{fig:MAE-train}}
\subfloat{\includegraphics[scale=0.48,clip]{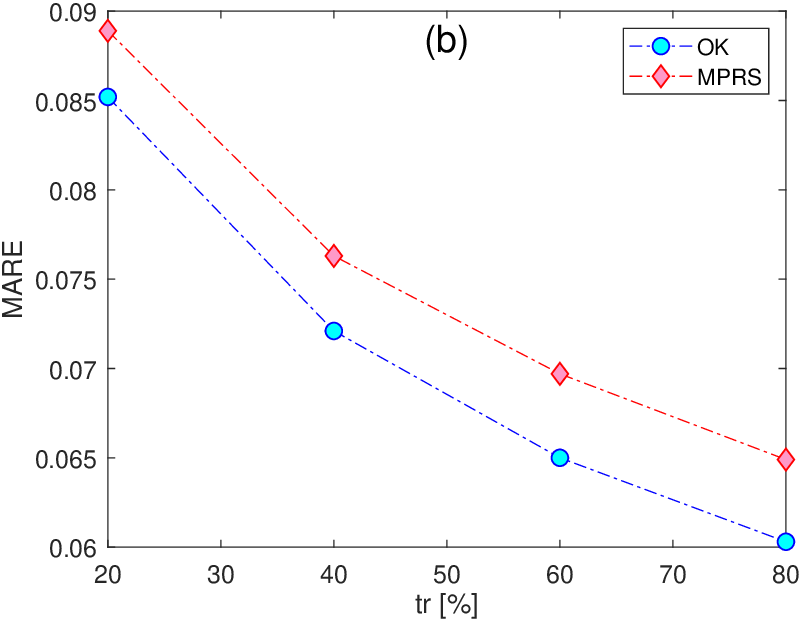}\label{fig:MARE-train}} \\
\subfloat{\includegraphics[scale=0.48,clip]{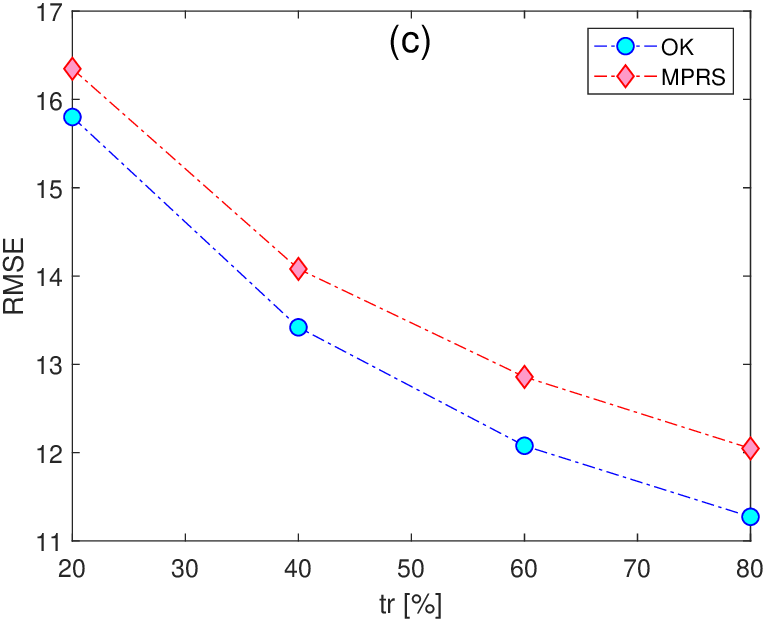}\label{fig:RMSE-train}} 
\subfloat{\includegraphics[scale=0.48,clip]{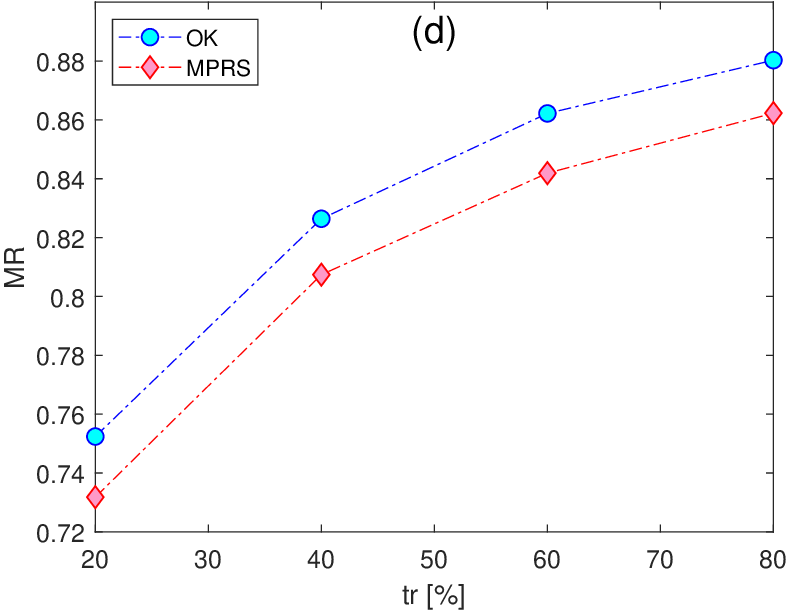}\label{fig:R-train}} \\
\subfloat{\includegraphics[scale=0.48,clip]{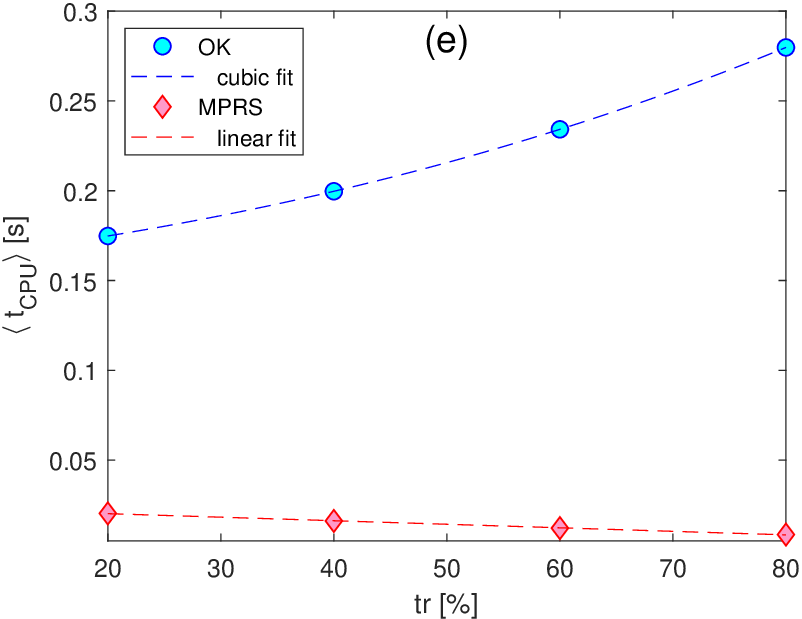}\label{fig:CPU-train}}
\caption{Dependence of  MPRS and OK validation measures on the ratio of training points $tr$. The measures are calculated from 100 realizations of a Gaussian random field with $m= 150$, $\sigma=25$ and covariance model WM($\kappa=0.2,\nu=0.5$); the field  is  sampled at $1,000$ scattered points inside a square domain of length $L=50$.}
  \label{fig:err-train}
\end{figure}

In Fig.~\ref{fig:err-train} we present the evolution of all the measures with increasing $tr$. As expected, for higher $tr$ values both methods give smaller errors and larger $MR$. The differences between MPRS and OK persist  and  seem to slightly increase with  increasing $tr$. While the relative prediction performance of MPRS slightly decreases its relative computational efficiency substantially increases---for $tr=0.8$ MPRS is 33 times faster than OK. The computational complexity of OK increases cubically with  sample size. In contrast, the MPRS computational cost only depends on $P$ ($\#$ of prediction points), which  decreases with increasing $tr$. The fit in Fig.~\ref{fig:CPU-train} indicates an approximately linear decrease.

\begin{figure}[t!]
\centering
\subfloat{\includegraphics[scale=0.48,clip]{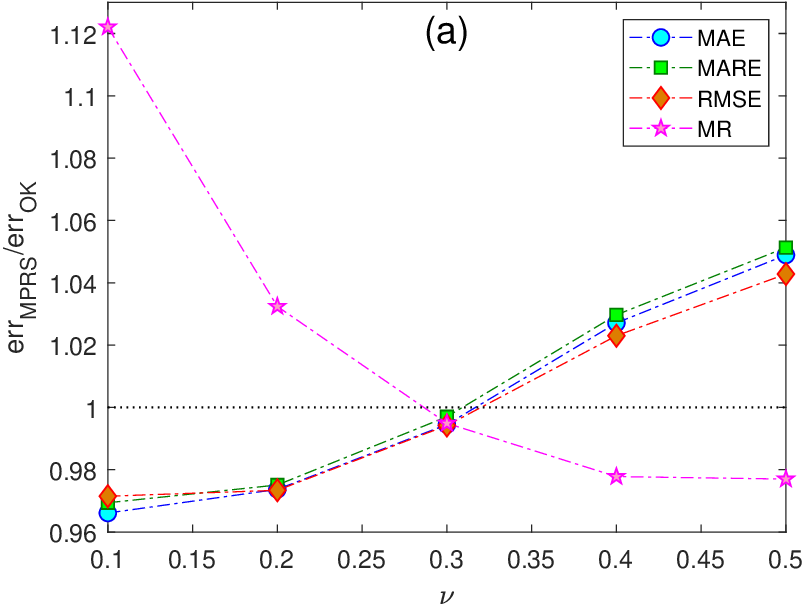}\label{fig:rel_err-nu_N_train33}}
\subfloat{\includegraphics[scale=0.48,clip]{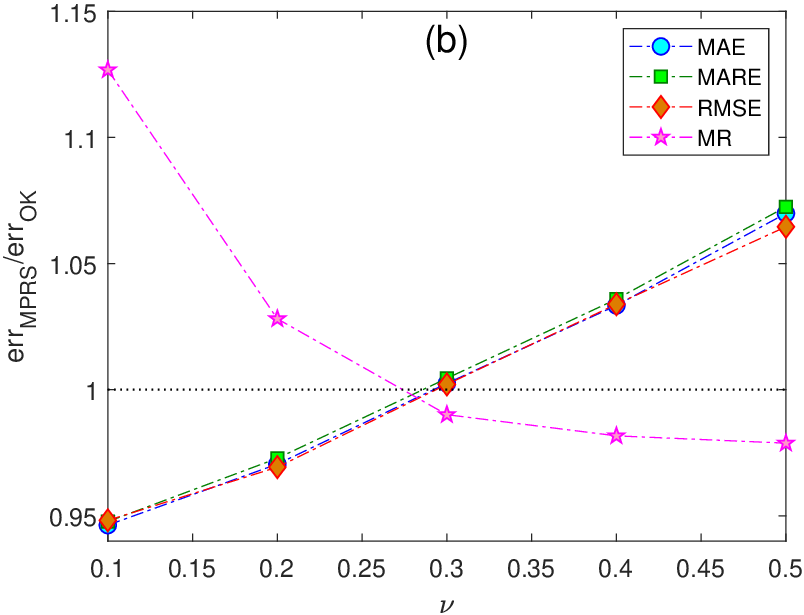}\label{fig:rel_err-nu_N_train66}}
\caption{Dependence of the ratios of MPRS and OK validation measures on the smoothness parameter. The measures are calculated based on 100 realizations of a Gaussian random field with $m= 150$, $\sigma=25$ and the covariance model WM($\kappa=0.2,\nu$), sampled at $1,000$ scattered points inside a square domain of length $L=50$. Panels (a) and (b) show the results for $tr=0.33$ and $0.66$, respectively.}
  \label{fig:rel_err-nu}
\end{figure}

Next, we evaluate the relative MPRS prediction performance with increasing data roughness, i.e.,   gradually decreasing  $\nu$. In Figs.~\ref{fig:rel_err-nu_N_train33} and~\ref{fig:rel_err-nu_N_train66} we present the ratios of different calculated measures (errors) obtained by MPRS and OK  for $tr=0.33$ and $0.66$, respectively. The plots exhibit a consistent decrease (increase) of MPRS errors (correlation coefficient) with decreasing smoothness  from $\nu=0.5$ to $0.1$. At $\nu \approx 0.3$ the MPRS and OK validation measures become approximately identical,  and for $\nu \lesssim 0.3$ MPRS outperforms OK. Thus, MPRS seems to be more appropriate than OK for the interpolation of rougher data.

\begin{figure}[t!]
\centering
\includegraphics[scale=0.5,clip]{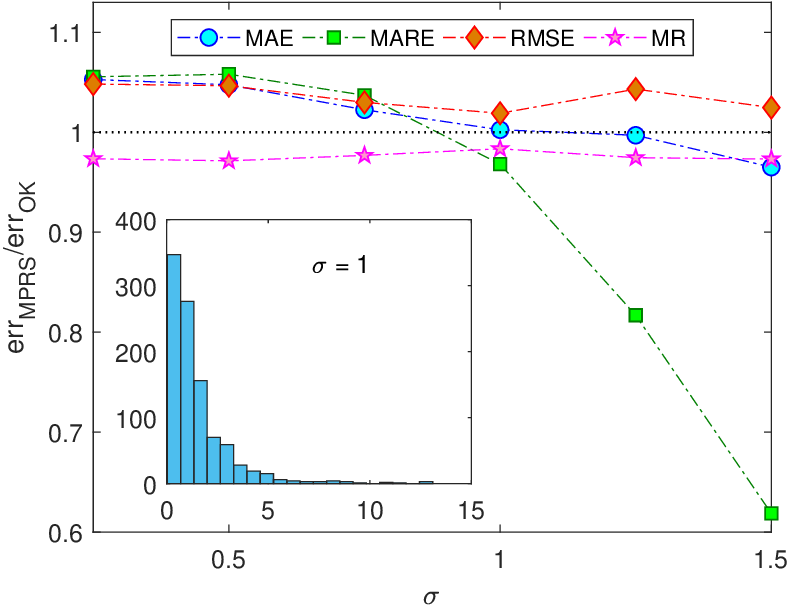}
\caption{Dependence of the ratios of  MPRS and OK validation measures on the random field skewness (controlled by  $\sigma$). The measures are calculated from 100 realizations of a lognormal random field with $m= 0$ and covariance model WM($\kappa=0.2,\nu=0.5$), sampled at $1,000$ scattered points inside a square domain of length $L=50$, for $tr=0.33$. The inset shows, as an example, the data distribution for $\sigma=1$.}
\label{fig:rel_err-sigma_log-N_train33}
\end{figure}

The above cases assume a Gaussian distribution, which is not universally observed in real-world data. To assess MPRS  performance for non-Gaussian (skewed) distributions, we simulate synthetic data that follow the lognormal law, i.e., $\log Z \sim N(m=0, \sigma)$ with  
the WM($\kappa=0.2,\nu=0.5$) covariance. The lognormal random field that generates the data has median $z_{0.50}= \exp(m) =1$ and standard deviation $\sigma_{Z}= \left[ \exp(\sigma^{2})-1 \right]^{1/2} \exp(m+ \sigma^{2}/2)$. Thus,  the parameter $\sigma$ controls the  data skewness (see the inset of Fig.~\ref{fig:rel_err-sigma_log-N_train33}). Figure~\ref{fig:rel_err-sigma_log-N_train33} demonstrates that MPRS can provide better interpolation performance compared to OK  for non-Gaussian, highly skewed data as well. In particular, for $\sigma \gtrsim 1$ the MAE and MARE measures of MPRS become comparable or smaller than those obtained with OK.  

\begin{figure}[t!]
\centering
\includegraphics[scale=0.5,clip]{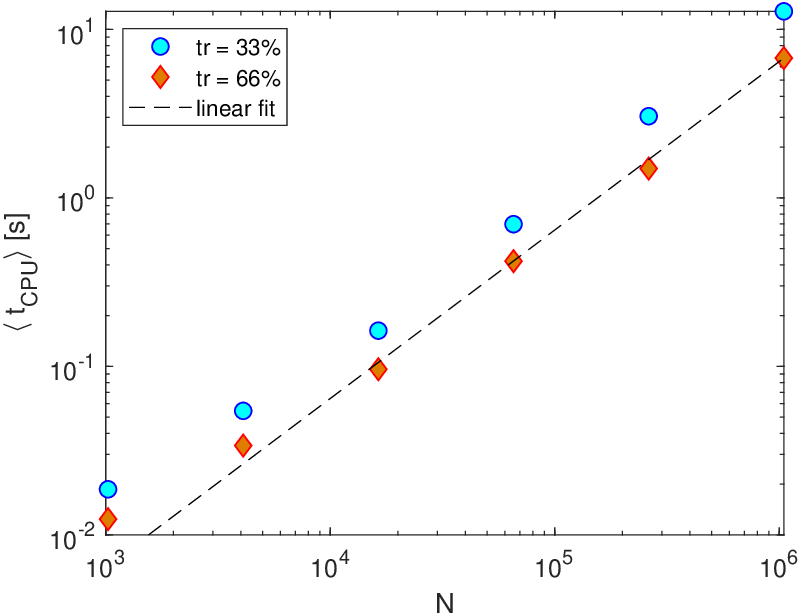}
\caption{CPU time scaling vs  data size $N$ based on 100 samples of Gaussian RFs with  covariance model WM($\kappa=0.2,\nu=0.5$). Two plots for $tr=0.33$ (squares) and $tr=0.66$ (diamonds) are shown. The dashed line is a guide to the eye for linear dependence.}
\label{fig:t_cpu}
\end{figure}

Finally, we assess the computational complexity of MPRS by measuring the CPU time necessary for interpolation of $\lceil (1-tr)\,N \rceil$ points based on $\lfloor tr\,N \rfloor$ samples,  for  increasing $N$ between $2^{10}$ and $2^{20}$. The results presented in Fig.~\ref{fig:t_cpu} for  $tr=0.33$ and $0.66$ confirm approximately linear (sublinear for smaller $N$) dependence, already suggested in Fig.~\ref{fig:CPU-train}. The CPU times obtained for $tr=0.66$, which involve more samples but fewer prediction points, are systematically smaller. 

\subsection{Real 2D spatial data}
\label{subsec:results_real_sp}
\subsubsection{Ambient gamma dose rates}
The first two data sets represent radioactivity levels in the routine and the simulated emergency scenarios~\citep{SIC2004}. In particular, the routine data set represents daily mean gamma dose rates over Germany reported by the national automatic monitoring network at  $1,008$ monitoring locations. In the second data set an accidental release of radioactivity in the environment was simulated in the South-Western corner of the monitored area. These data were used in Spatial Interpolation Comparison (SIC) 2004 exercise  to test the prediction performance of various methods. 

The training set involved daily data from 200 randomly selected stations, while the validation set involved the remaining 808 stations.  Data summary statistics and an extensive discussion of different interpolation approaches are found in~\citet{SIC2004}. In total, 31 algorithms were applied  to the routine scenario while several geostatistical techniques failed in the emergency scenario due to instabilities caused by the outliers (simulated release data).

The validation measures from MPRS and OK  applied to both the routine and emergency data sets are presented in Table~\ref{table:SIC2004}. The OK code that we used  failed to fit the spatial dependence in the routine data; thus, we show the average measures of the two OK approaches reported in~\citet{SIC2004}. 
Comparing the results for OK and MPRS, for the routine data set OK gives slightly better results than MPRS. However, for the emergency data set AE and ARE errors are much smaller for MPRS, while OK gives superior values for the RSE and $R$ validation measures.

\begin{table}[t!]
\caption{Interpolation validation measures (obtained from Table~\ref{tab:val-measures} for $V=1$) for the MPRS and OK methods applied to the routine and emergency SIC2004 data sets. OK* results for the routine data set are taken from~\citet{SIC2004}.} 
\label{table:SIC2004}
\centering
\begin{tabular}{lcccccc}
\hline
\hfil Data & \hfil Method & \hfil AE & \hfil ARE (\%) & \hfil RSE & \hfil $R$ (\%) & \hfil $t_{\mathrm{cpu}}$ (s)  \\ \hline
 Routine & MPRS & $9.21$ &  $ 9.28 $ & $ 12.58 $& $ 78.08$&  $ 0.02 $\\
        & OK*   & $9.11$ &  $ - $ & $ 12.47 $& $ 78.50$&  $ - $\\ 
\midrule
 Emergency & MPRS & $18.57$ &  $ 12.63 $ & $ 76.85 $& $ 40.62$&    $ 0.02 $\\
        & OK   & $22.05$ &  $ 18.74 $ & $ 74.19 $& $ 47.05$&    $ 0.35 $\\
\hline
\end{tabular}
\end{table}

\begin{figure}[t!]
\centering
\includegraphics[scale=0.5,clip]{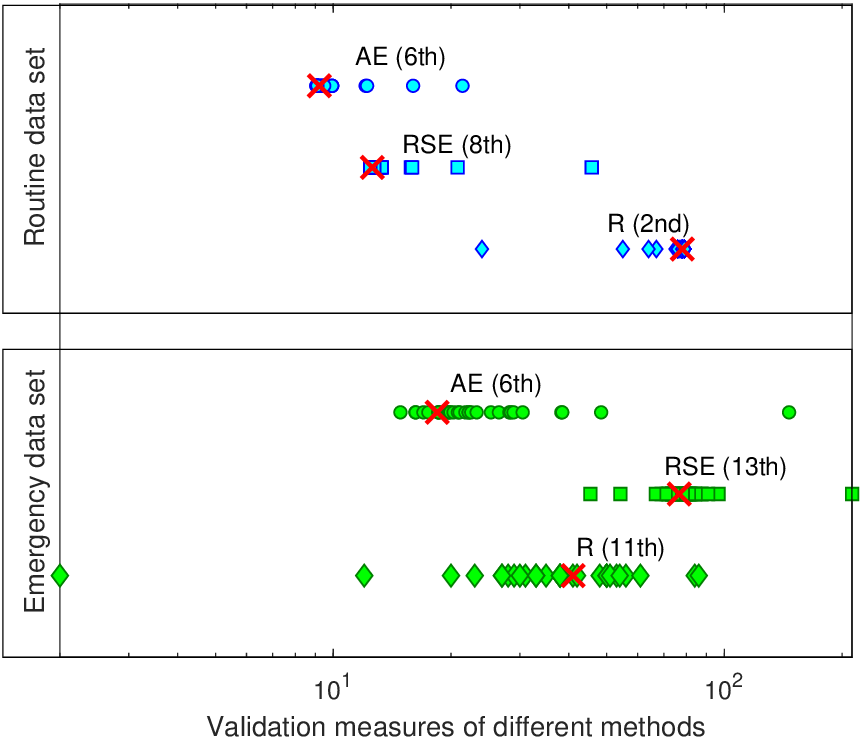}
\caption{Values of AE, RSE and $R$ obtained for the routine and emergency data sets by means of 31 interpolation methods reported in the SIC2004 exercise (circles, squares and diamonds) and the MPRS approach (red crosses). The numbers in parentheses denote the ranking of MPRS performance for the particular validation measure with respect to all 32 methods.  }
\label{fig:SIC04}
\end{figure}

In Fig.~\ref{fig:SIC04} we compare the MPRS performance with the   results obtained with the 31 different approaches reported in~\citet{SIC2004}. This comparison shows that MPRS is  competitive with  geostatistical,  neural network, support vector machines and splines. In particular, for the routine data set MPRS ranked 6th, 8th and 2nd  for AE, RSE and $R$, respectively,  and for the emergency data 6th, 13th and 11th.

\subsubsection{Jura data set}

\begin{figure}[b!]
\centering
\includegraphics[scale=0.5,clip]{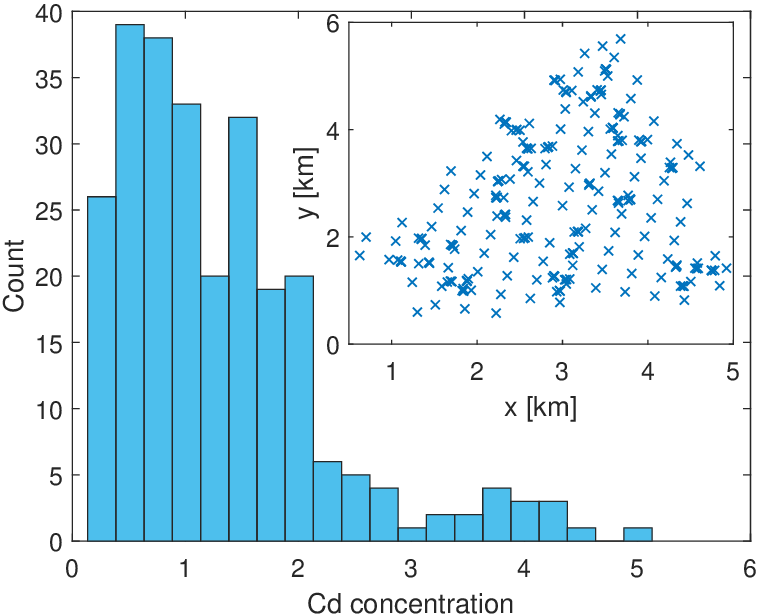}
\caption{Histogram of cadmium soil contamination concentrations from the Jura data set. The inset shows the measurement locations.}
\label{fig:Jura_Cd_hist}
\end{figure}

This data set comprises  topsoil heavy metal concentrations (in ppm) in the Jura Mountains (Switzerland)~\citep{atteia94,goov97}.  In particular, the data set includes concentrations of the following metals: Cd, Co, Cr, Cu, Ni, Pb, Zn. The 259 measurement locations and the histogram of Cd concentrations, as an example, are shown in Fig.~\ref{fig:Jura_Cd_hist}. The detailed statistical summary of all the data sets can be found in~\citet{atteia94,goov97}.

\begin{figure}[t!]
\centering
\subfloat{\includegraphics[scale=0.48,clip]{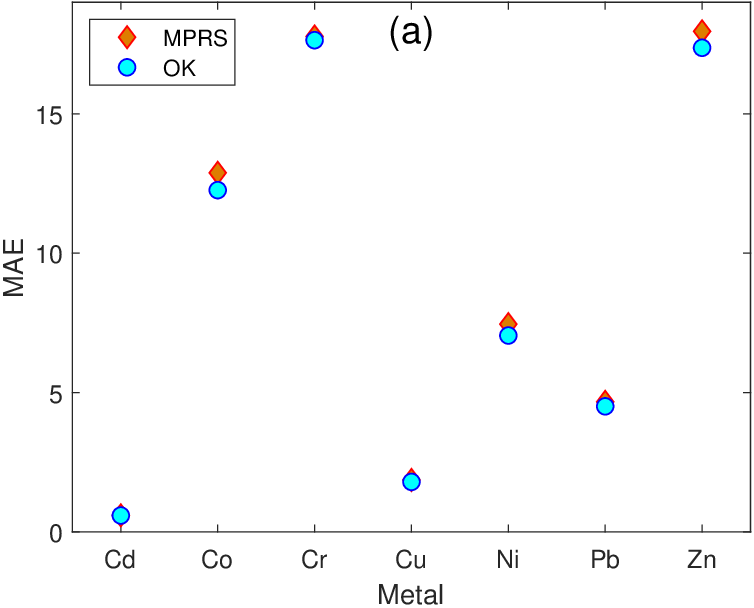}\label{fig:MAE-Jura}}
\subfloat{\includegraphics[scale=0.48,clip]{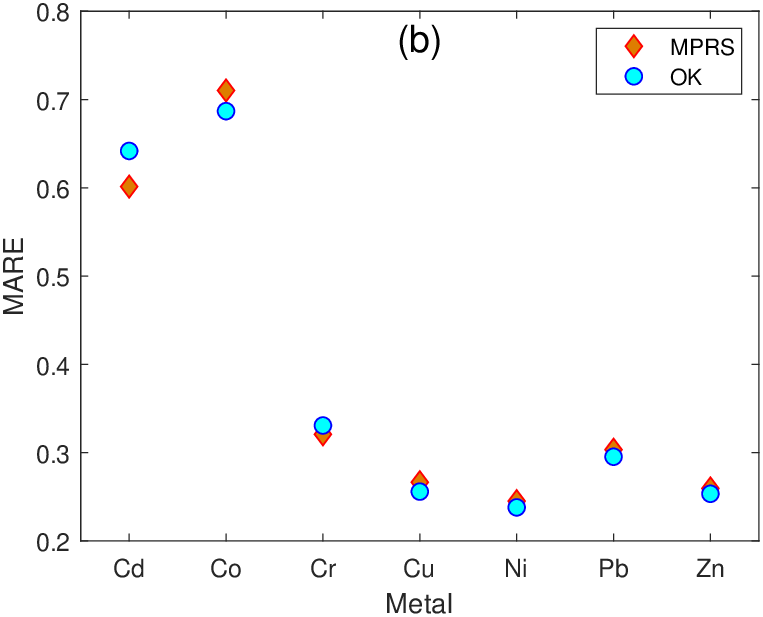}\label{fig:MARE-Jura}} \\
\subfloat{\includegraphics[scale=0.48,clip]{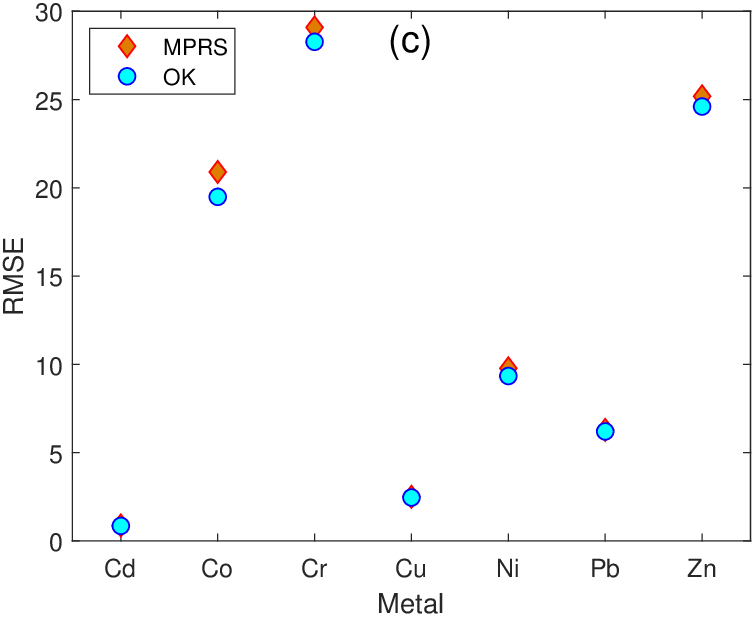}\label{fig:RMSE-Jura}} 
\subfloat{\includegraphics[scale=0.48,clip]{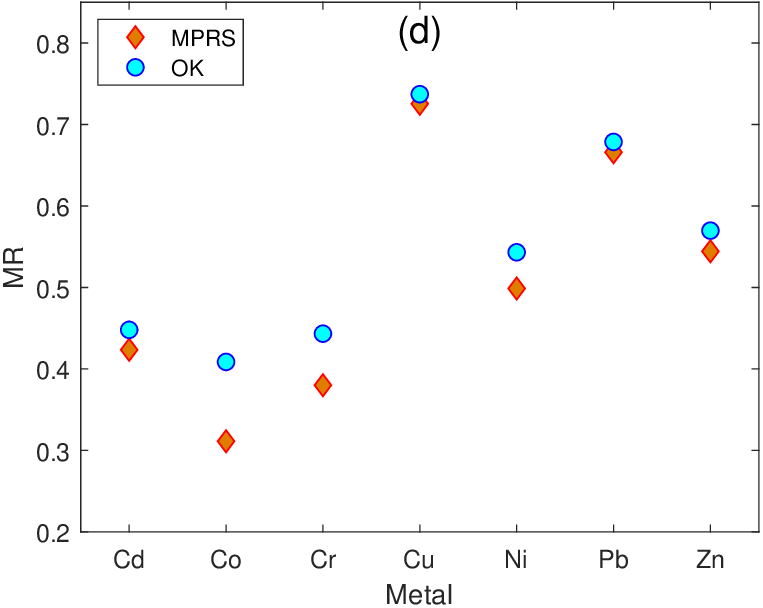}\label{fig:R-Jura}} \\
\subfloat{\includegraphics[scale=0.48,clip]{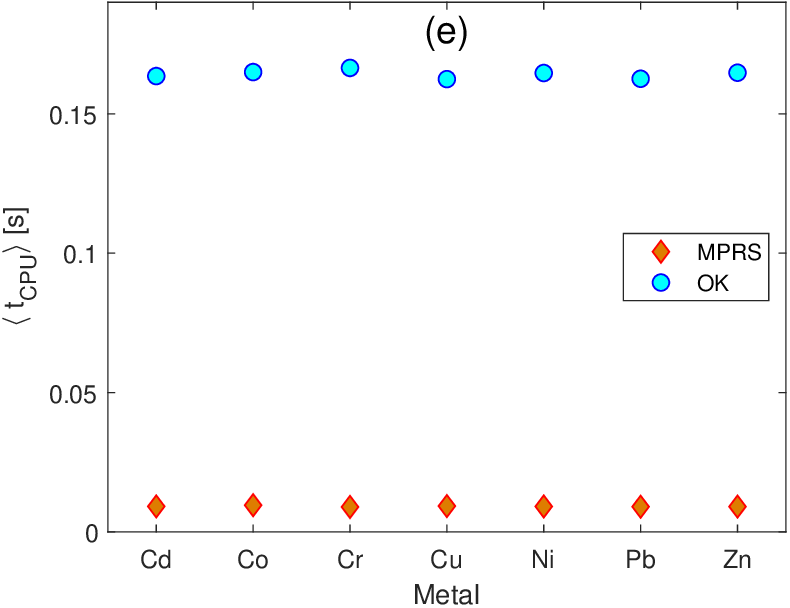}\label{fig:CPU-Jura}}
\caption{MPRS and OK validation measures for the Jura heavy metal contamination data set. The measures are calculated from 100 randomly chosen training sets of 85 points leaving 174 points in the validation sets.}
  \label{fig:err-jura}
\end{figure}

\begin{figure}[t!]
\centering
\includegraphics[scale=0.5,clip]{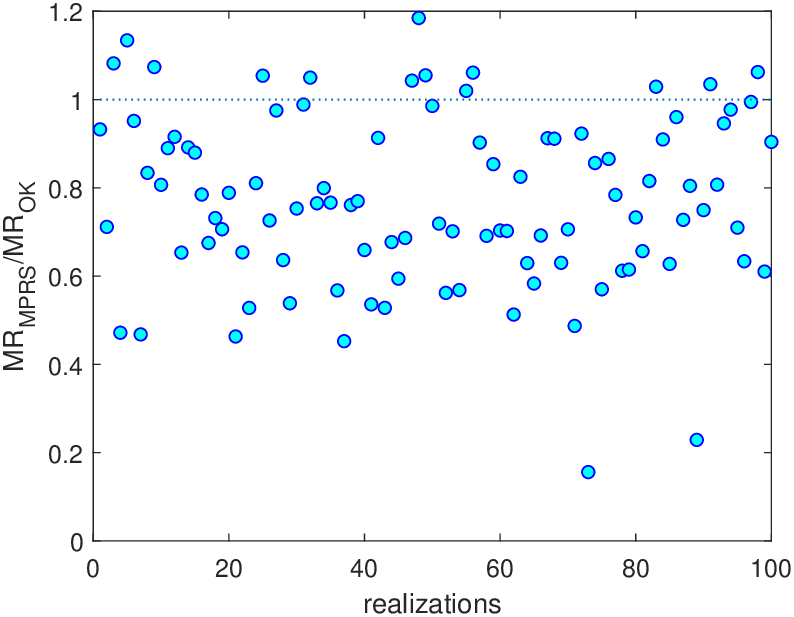}
\caption{Ratio of mean correlation coefficients, $MR_{\rm{MPRS}}/MR_{\rm{OK}}$, between validation data and predicted values  for Co concentration. The mean is evaluated over 100 random validation splits.}
\label{fig:R_MPRS_vs_OK}
\end{figure}

For each data set we generate $V=100$ different training sets consisting of 85 randomly selected points. Different panels in Fig.~\ref{fig:err-jura} compare MPRS and OK  validation measures for the 7 metal concentrations. In most cases OK gives smaller (larger) errors ($MR$) than MPRS. However, MPRS gives lower MARE values for Cd and Cr concentrations. The differences between MPRS and OK errors are on the order of a few percent. The largest differences appear for mean $R$: the maximum relative difference,   reaching over $20\%$, was recorded for Co. Nevertheless, due to relatively large sample-to-sample fluctuations even in such cases, for certain splits MPRS shows better performance than OK. Fig.~\ref{fig:R_MPRS_vs_OK}  shows the $R$ ratios per split for the two methods. In 13 instances MPRS gives larger values than OK. The execution times, presented in Fig.~\ref{fig:CPU-Jura}, demonstrate that MPRS is about 18 times faster than OK.

\subsubsection{Walker lake data set}
\begin{figure}[t!]
\centering
    \subfloat{\includegraphics[scale=0.5,clip]{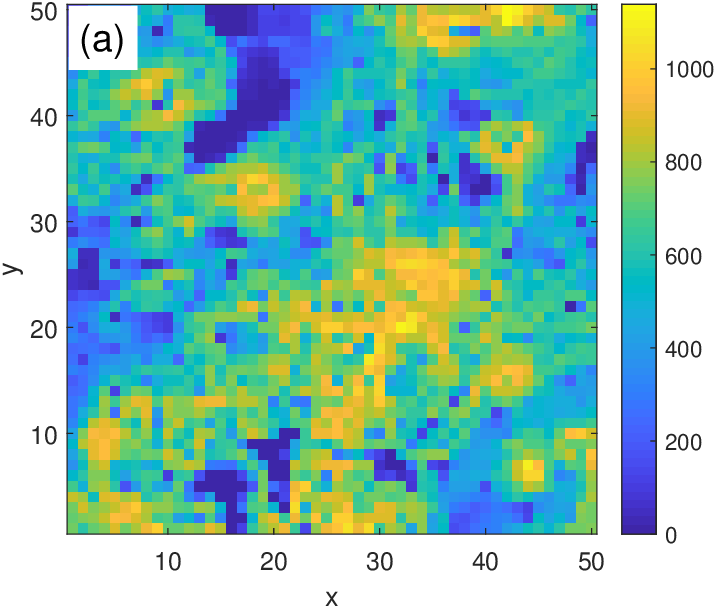}\label{fig:zo_walker}}
    \subfloat{\includegraphics[scale=0.5,clip]{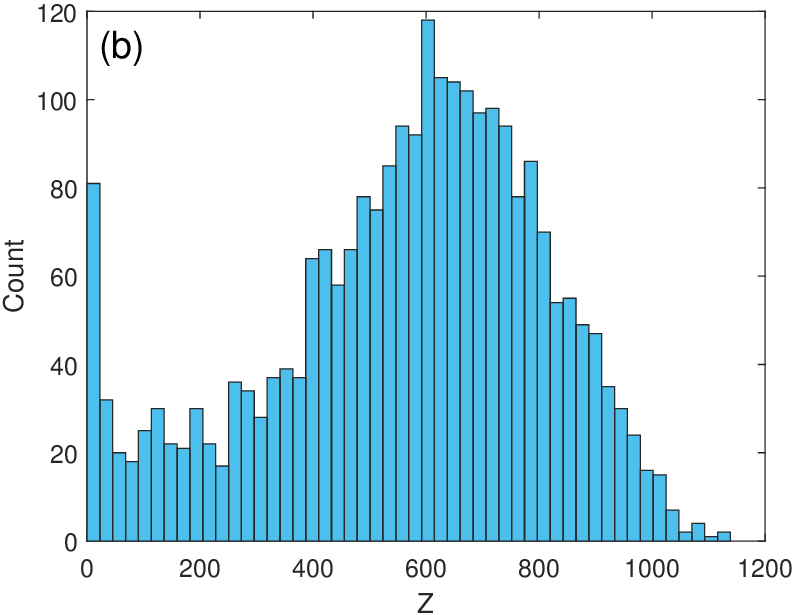}\label{fig:hist_walker}}
    \caption{Walker lake data (a) spatial distribution and (b) histogram.}
  \label{fig:walker_lake}
\end{figure}

This data set demonstrates the ability of MPRS to fill data gaps  on  rectangular grids. The  data represent  DEM-based chemical concentrations with units in parts per million (ppm) from the Walker lake area in Nevada~\citep{isaaks89}. We use a subset of the full grid comprising a square of  size $L \times L$ with $L = 50$. The summary statistics are: $N=2,500$, $z_{\min}=0$, $z_{\max}= 1,138.6$, $\bar{z}=564.14$, $z_{0.50}= 601.10$, $\sigma_{z}=245.77$, \emph{skewness} ($s_z$) and \emph{kurtosis} ($k_z$) coefficients  $s_z=-0.53$ and $k_z = 2.74$, respectively. The spatial distribution and histogram of the data are shown in Fig.~\ref{fig:walker_lake}.

\begin{table}[b!]
\caption{Interpolation validation measures for MPRS and OK. The measures are based on 100 randomly chosen training sets that include $tr\, N$ of the  $N=2,500$ points (Walker lake data set).} 
\label{table:walker}
\centering
\begin{tabular}{lccccc}
\hline
\hfil $tr$ & \hfil Method & \hfil MAE & \hfil RMSE & \hfil $MR$ (\%) & \hfil $\langle t_{\mathrm{cpu}} \rangle$ (s)  \\ \hline
 $0.33$ & MPRS & $117.74$ & $ 159.78 $&   $ 76.07$&    $ 0.03 $\\
  & OK & $114.82$ & $ 157.54$&  $ 77.05$&    $ 0.31 $\\ \noalign{\smallskip}
 $0.66$ & MPRS & $105.21$ & $ 143.90 $&   $ 81.03$&    $ 0.02 $\\
  & OK & $101.72$ & $ 140.41$&  $ 82.07$&    $ 1.03 $\\
\hline
\end{tabular}
\end{table}

\begin{figure}[t!]
\centering
\subfloat{\includegraphics[scale=0.48,clip]{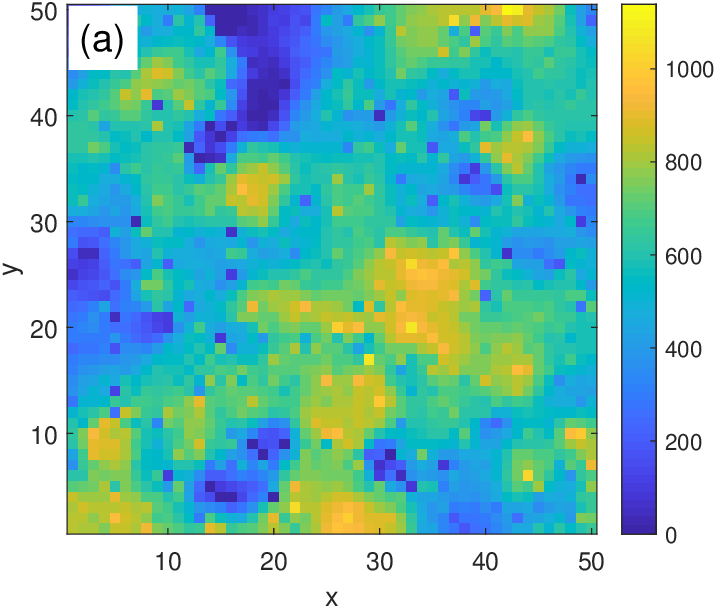}\label{fig:walker_lake_zr_mprs}}
\subfloat{\includegraphics[scale=0.48,clip]{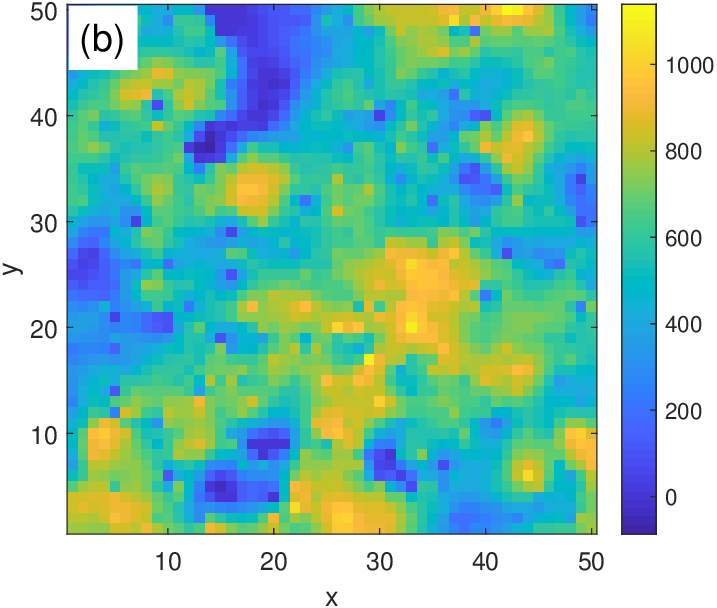}\label{fig:walker_lake_zr_ok}} \\
\subfloat{\includegraphics[scale=0.48,clip]{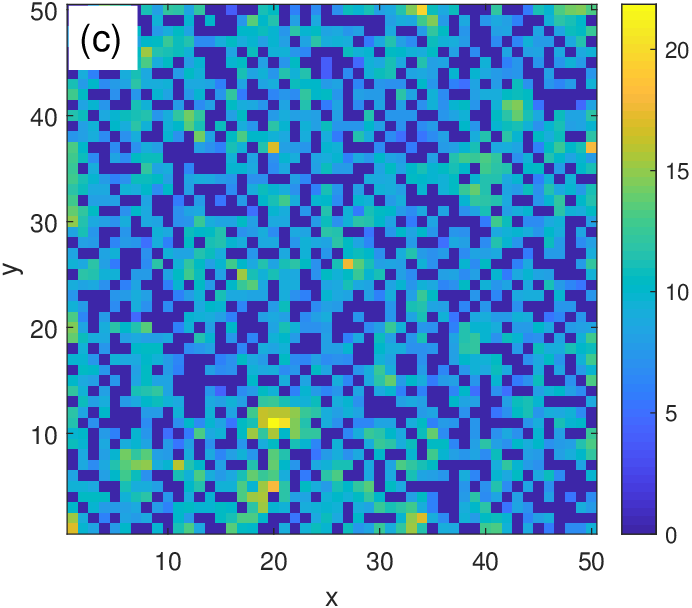}\label{fig:walker_lake_std_mprs}}
\subfloat{\includegraphics[scale=0.48,clip]{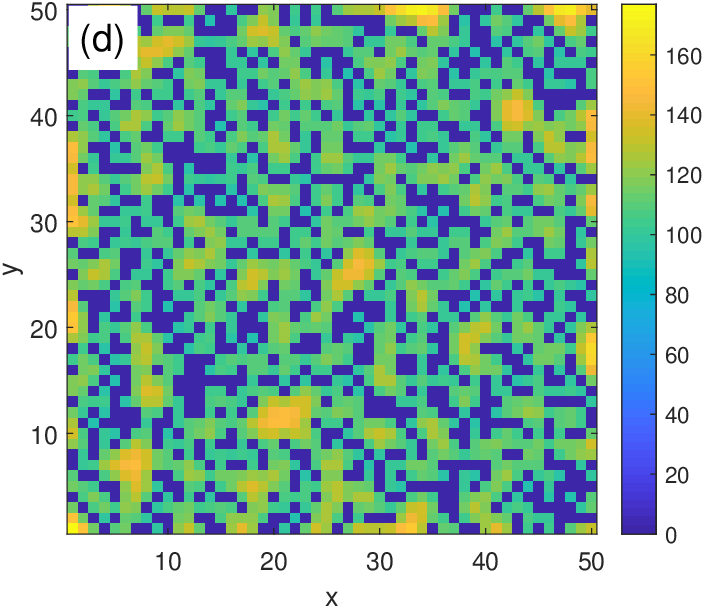}\label{fig:walker_lake_std_ok}}
\caption{Visual comparison of (a) MPRS and (b) OK interpolated maps and the corresponding prediction standard deviations, shown below in (c) and (d), respectively, for the Walker lake data with $tr=33\%$. The points with zero standard deviation predominantly coincide with the sample locations.}
  \label{fig:err-walker}
\end{figure}

Training sets of  size $\lfloor tr\,N \rfloor$ are generated  by removing randomly $\lceil (1-tr)\,N \rceil$ points from the full data set. For  $tr=0.33$ and $0.66$ we generate $V=100$ different training-validation splits. The validation measures are listed in Table~\ref{table:walker}. Due to zero values, the relative MARE errors can not be evaluated. The results are consistent with the previous tests in terms of slightly larger (smaller) values of the MPRS errors ($MR$) and considerably faster execution times than those obtained by OK. Visual comparison of the reconstructed gaps, shown in Fig.~\ref{fig:err-walker}, shows that the OK generated map is smoother than the MPRS reconstructed map, while the MPRS uncertainty estimates are considerably smaller than the OK estimates. We recall that the   MPRS prediction variance  is obtained from $M$ equilibrium realizations; thus, its small magnitude is attributed to the fact that at very low temperatures (e.g., the default value $T \approx 10^{-3}$), large fluctuations are strongly suppressed.

\subsubsection{Atmospheric latent heat release}
\begin{figure}[t!]
\centering
\subfloat{\includegraphics[scale=0.5,clip]{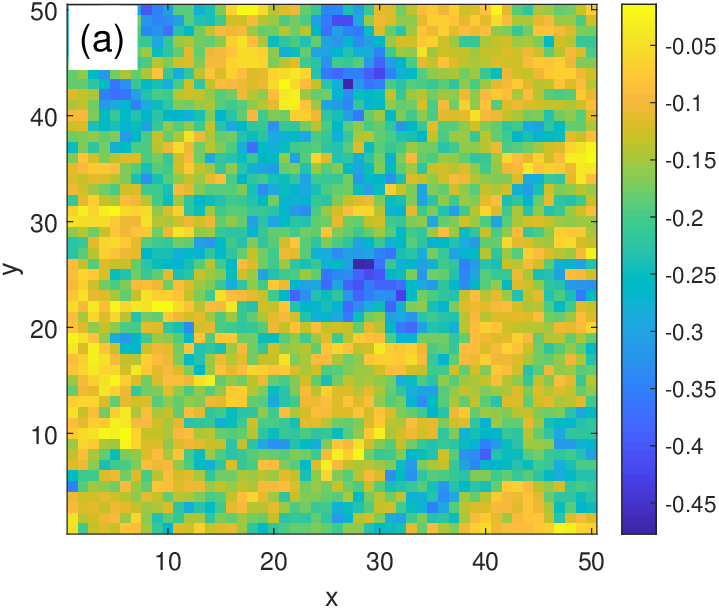}\label{fig:zo_heat}}    
\subfloat{\includegraphics[scale=0.5,clip]{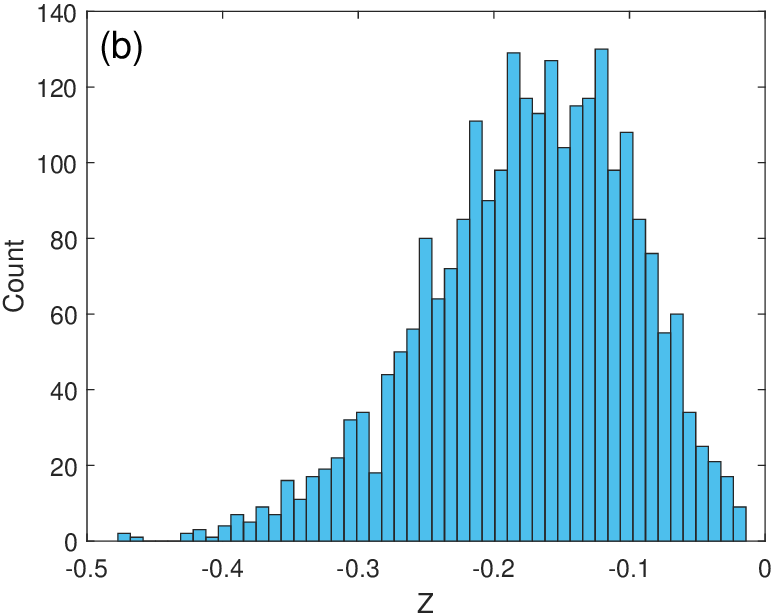}\label{fig:hist_heat}}
\caption{Monthly averaged latent heat release data measured in degrees Celsius per hour; (a) spatial distribution and (b) histogram.}
  \label{fig:latent_heat}
\end{figure}

This section focuses on monthly (January 2006) means of vertically averaged atmospheric latent heat release (measured in degrees Celsius per hour) measurements~\citep{Tao06,{TRMM11}}. The $L \times L$ data grid  ($L = 50$) extends in latitude from 16S to 8.5N and in longitude from 126.5E to 151E with cell size $0.5^\circ \times 0.5^\circ$. This area is in the Pacific region and extends over the Eastern part of the Indonesian archipelago. The data  summary statistics are as follows: $N=2,500$, $z_{\min}=-0.477$, $z_{\max}= -0.014$, $\bar{z}=-0.174$, $z_{0.50}= -0.168$, $\sigma_{z}=0.076$, $s_{z} = -0.515$, and  $k_{z}=3.122$. Negative (positive) values correspond to latent heat absorption (release). The spatial distribution and histogram of the data are shown in Fig.~\ref{fig:latent_heat}.

\begin{figure}[t!]
\centering
\subfloat{\includegraphics[scale=0.48,clip]{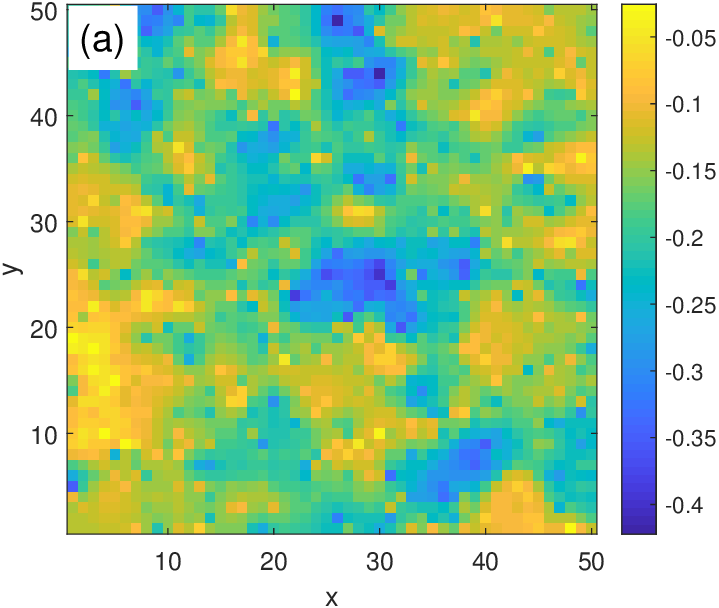}\label{fig:latent_zr_mprs}}
\subfloat{\includegraphics[scale=0.48,clip]{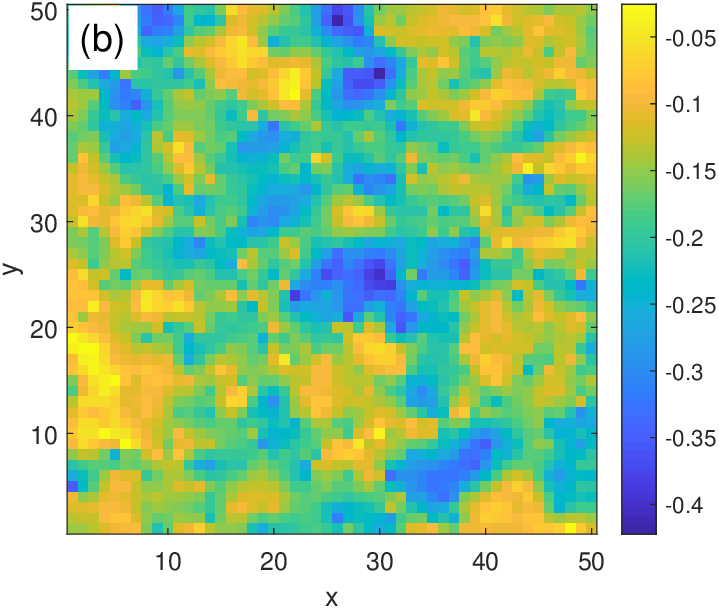}\label{fig:latent_zr_ok}} \\
\subfloat{\includegraphics[scale=0.48,clip]{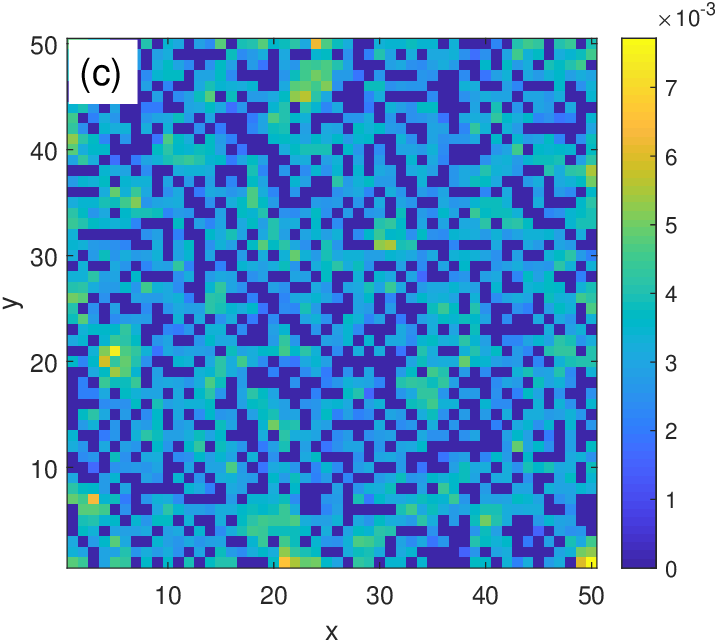}\label{fig:latent_std_mprs}}
\subfloat{\includegraphics[scale=0.48,clip]{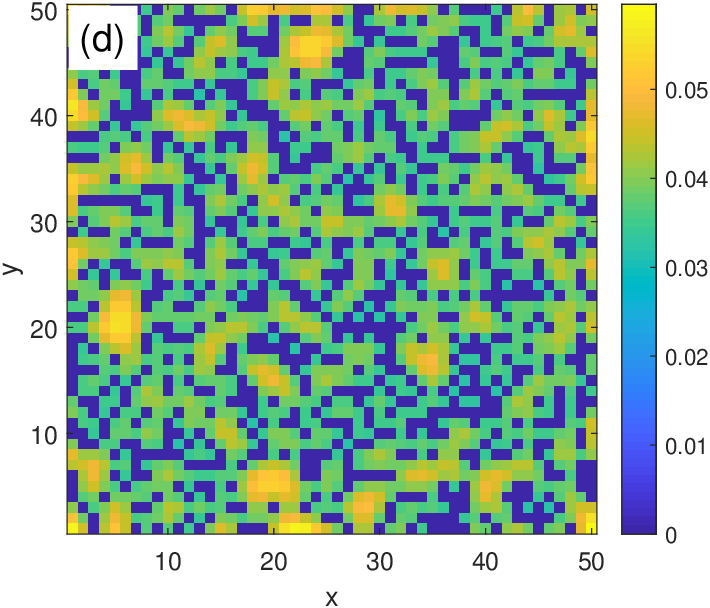}\label{fig:latent_std_ok}}
\caption{Visual comparison of (a) MPRS and (b) OK  interpolated maps (latent heat data) and  corresponding prediction standard deviations (c,d) for a $tr=0.33$ training-validation split.}
  \label{fig:err-latent}
\end{figure}

The comparison of validation measures  presented in Table~\ref{table:latent} and a visual comparison of the reconstructed maps and prediction uncertainty, shown in Fig.~\ref{fig:err-latent}, reveals similar patterns as  the Walker lake data: MPRS displays somewhat worse prediction performance but is significantly more efficient computationally than OK.

\begin{table}[h!]
\caption{Interpolation validation measures for  MPRS and OK  based on 100 randomly chosen training-validation splits; the training set includes $tr N$ points where $N=2,500$ (latent heat data set).} 
\label{table:latent}
\begin{tabular}{lcccccc}
\hline
\hfil $tr$ & \hfil Method & \hfil MAE  & \hfil MARE (\%) & \hfil RMSE & \hfil $MR$ (\%) & \hfil $\langle t_{\mathrm{cpu}} \rangle$ (s)  \\ \hline
 $33\%$ & MPRS & $0.042$ &   $ -32.92 $ & $ 0.053 $& $ 70.94$&    $ 0.03 $\\
        & OK   & $0.041$ &   $ -31.27 $ & $ 0.052 $& $ 72.58$&    $ 0.31 $\\ \noalign{\smallskip}
 $66\%$ & MPRS & $0.038$ &   $ -29.54 $ & $ 0.048 $& $ 77.13$&    $ 0.02 $\\
        & OK   & $0.036$ &   $ -27.22 $ & $ 0.046 $& $ 79.11$&    $ 1.02 $\\
\hline
\end{tabular}
\end{table}

\subsection{Time series (temperature and precipitation)}
\label{subsec:results_real_ts}
The MPRS method can be applied to data in any dimension $d$. We demonstrate that the MPRS method provides competitive predictive and computational performance for time series as well. 

\begin{figure}[t!]
\centering
\subfloat{\includegraphics[scale=0.48,clip]{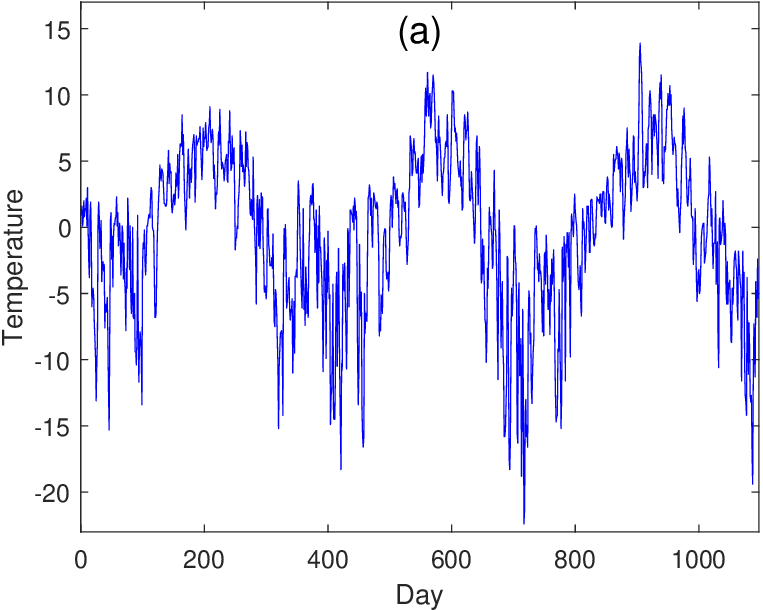}\label{fig:daily_temp_ts}}
\subfloat{\includegraphics[scale=0.48,clip]{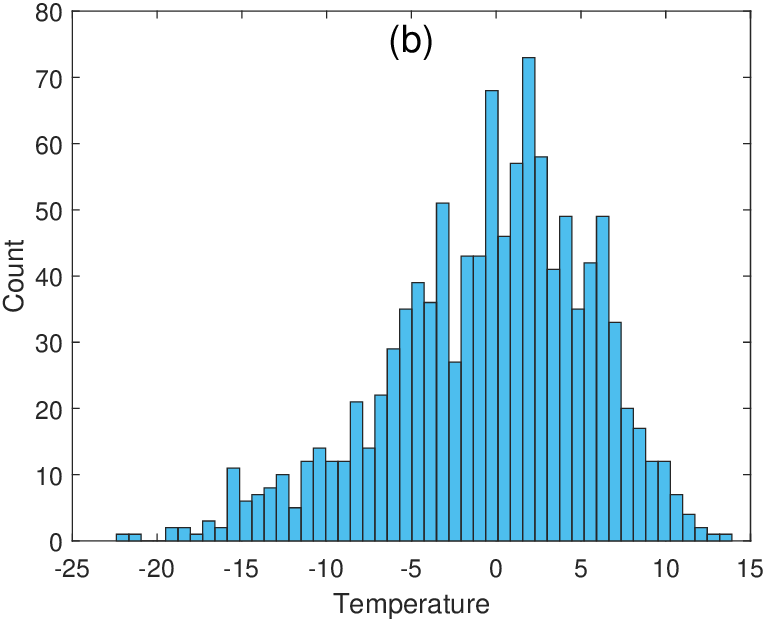}\label{fig:daily_temp_hist}} \\
\subfloat{\includegraphics[scale=0.48,clip]{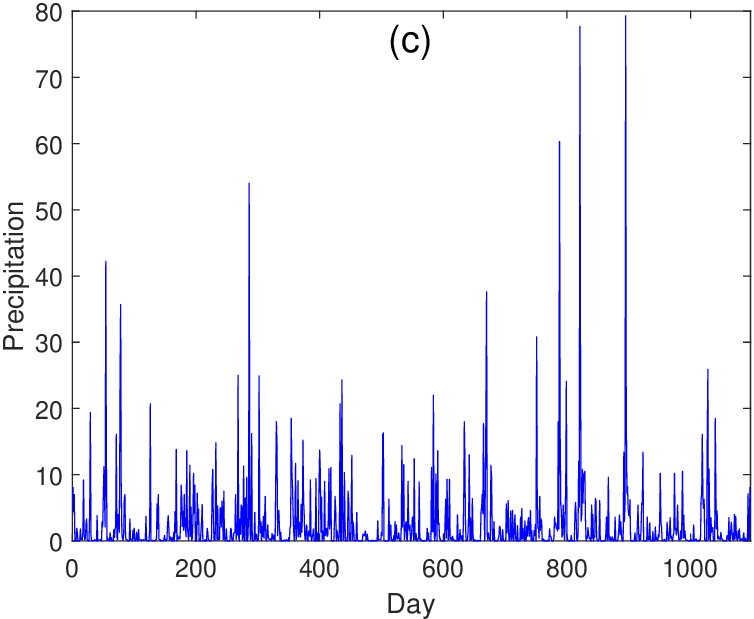}\label{fig:daily_prec_ts}}
\subfloat{\includegraphics[scale=0.48,clip]{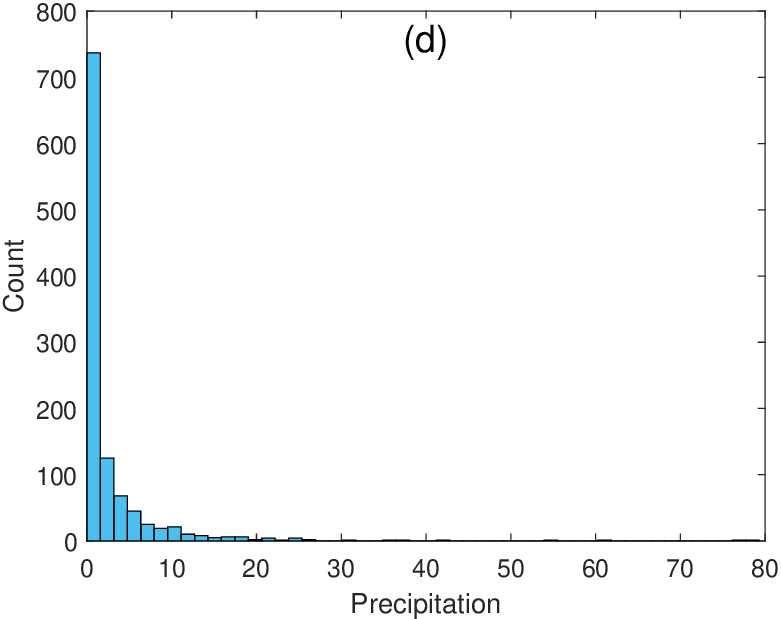}\label{fig:daily_prec_hist}}
\caption{Time series (a,c)  and respective frequency histograms (b,d) of daily temperature  (a,b) and  precipitation (c,d) at J\"{o}kulsa Eystri River (Iceland) for the period between January 1, 1972 and December 31, 1974.}
  \label{fig:ts}
\end{figure}

We consider two time series of daily data at J\"{o}kulsa Eystri River (Iceland), collected at the Hveravellir meteorological station, for the period between January 1, 1972 and December 31, 1974 (a total of $N=1,096$ observations)~\citep{tong90}. The first set represents daily temperatures (in degrees Celsius) and the second daily precipitation (in millimeters). The time series and the respective histograms are shown in Fig.~\ref{fig:ts}. The summary statistics for temperature are: $z_{\min}=-22.4$, $z_{\max}= 13.9$, $\bar{z}=-0.441$, $z_{0.50}= 0.3$, $\sigma_{z}=6.021$,  $s_{z}=-0.595$, and $k_{z}=3.196$ . The precipitation statistics are: $z_{\min}=0$, $z_{\max}= 79.3$, $\bar{z}=2.519$, $z_{0.50}= 0.3$, $\sigma_{z}=6.025$, $s_{z}=6.512$, and $k_{z}=65.268$. The temperature follows an almost Gaussian distribution, while precipitation is strongly non-Gaussian, highly skewed, with the majority of values equal or close to zero and a small number of outliers that form an extended right tail.

The interpolation validation measures and computational times for  MPRS and OK are listed in Table~\ref{table:ts}. The results are based on 100 randomly selected training-validation splits which include $tr N$ points. For the temperature data, the MPRS performance  relative to OK  is similar as for the 2D spatial data. 
However,  in the case of precipitation  MPRS returns a lower MAE than OK for $tr=0.33$, while  for $tr=0.66$ MPRS is clearly better for all measures. This observation agrees with the results for the synthetic spatial data, i.e.,  the relative performance of MPRS improves for strongly non-Gaussian data (cf. Fig.~\ref{fig:rel_err-sigma_log-N_train33} which displays relative errors for lognormal data with gradually increasing $s_{z}$).

\begin{table}[t!]
\caption{Interpolation validation measures for MPRS and OK  based on 100 randomly selected training-validation splits. The training sets include $tr N$ points ($N=1096$ and $tr=0.33, 0.66$) from the daily temperature and precipitation time series.} 
\label{table:ts}
\centering
\begin{tabular}{lcccccc}
\hline
\hfil Data & \hfil $tr$ & \hfil Method & \hfil MAE & \hfil RMSE & \hfil $MR$ (\%) & \hfil $\langle t_{\mathrm{cpu}} \rangle$ (s)  \\ \hline
Temperature & $33\%$ & MPRS & $2.2726$ & $ 3.1404 $&   $ 85.73$&    $ 0.02 $\\
 & & OK & $2.1896$ & $ 3.0413$&  $ 85.48$&    $ 0.35 $\\ \noalign{\smallskip}
& $66\%$ & MPRS & $1.7639$ & $ 2.4676 $&   $ 91.09$&    $ 0.01 $\\
&  & OK & $1.6635$ & $ 2.3568$&  $ 91.94$&    $ 0.40 $\\ \noalign{\smallskip}
Precipitation & $33\%$ & MPRS & $3.1357$ & $ 6.8961 $&   $ 17.52$&    $ 0.02 $\\
 & & OK & $3.2661$ & $ 6.7884$&  $ 18.98$&    $ 0.34 $\\ \noalign{\smallskip}
& $66\%$ & MPRS & $2.8028$ & $ 6.2041 $&   $ 27.92$&    $ 0.01 $\\
&  & OK & $3.0604$ & $ 6.3194$&  $ 25.59$&    $ 0.41 $\\
\hline
\end{tabular}
\end{table}

\subsection{Real 3D spatial data}
\label{subsec:results_real_3d}
Finally, we study calcium and magnesium soil content  sampled in the 0-20cm soil layer~\citep{diggle07}.  There are $N=178$ observations and the data are measured in $mmol_c/dm^3$. The calcium data statistics are $z_{\min}=21$, $z_{\max}= 78$, $\bar{z}=50.68$, $z_{0.50}= 50.5$, $\sigma_{z}=11.08$, $s_{z}=-0.097$, and  $k_{z}=2.64$, while for magnesium the respective statistics are $z_{\min}=11$, $z_{\max}= 46$, $\bar{z}=27.34$, $z_{0.50}= 27$, $\sigma_{z}=6.28$,  $s_{z}=0.031$, and  $k_{z}=2.744$. 


In this case we compare MPRS with the IDW method~\citep{shepard1968two} using a  power exponent equal to 2 and unrestricted search radius. As evidenced in the validation measures (Table~\ref{table:3d}),  MPRS outperforms IDW in terms of prediction accuracy. The relative differences change from a few percent for $tr=0.33$  to $~\sim 12\%$ for $tr=0.66$. For this particular data set, IDW is computationally more efficient than MPRS.  
However, this is due to the limited data size. With increasing $N$ the relative computational efficiency of MPRS will improve and eventually outperform IDW, since the computational time for the former scales as ${\mathcal O}(P)$, while for the latter as  ${\mathcal O}(P\,N)$ [e.g., see  comparison of MPR and IDW~\citep{mz-dth18}].

\begin{table}[t!]
\caption{Interpolation validation measures for the MPRS and OK methods based on 100 randomly chosen training sets including $tr\%$ of the total number $N=178$ of points in calcium and magnesium contents in soil samples at the 0-20cm soil layer.} 
\label{table:3d}
\begin{tabular}{lccccccc}
\hline
\hfil Data & \hfil $tr$ & \hfil Method & \hfil MAE & \hfil MARE (\%) & \hfil RMSE & \hfil $MR$ (\%) & \hfil $\langle t_{\mathrm{cpu}} \rangle$ (s)  \\ \hline
Ca & $33\%$ & MPRS & $6.93$ & $ 14.67 $& $ 8.95 $&   $ 60.51$&    $ 0.006 $\\
                 &        & IDW & $7.16$ & $ 15.60$ &  $ 9.13$& $ 59.71$&  $ 0.001 $\\ \noalign{\smallskip}
                 & $66\%$ & MPRS & $6.43$ & $ 13.60 $ & $ 8.38 $&   $ 65.49$&    $ 0.005 $\\
                 &        & IDW & $6.92$ & $ 15.15 $& $ 8.82$&  $ 64.35$&    $ 0.001 $\\ \noalign{\smallskip}
Mg & $33\%$ & MPRS & $4.31$ & $ 17.56 $ & $ 5.46 $&   $ 52.40$&    $ 0.006 $\\
                  &        & IDW & $4.48$ & $ 18.54 $ & $ 5.59$&  $ 49.68$&    $ 0.001 $\\ \noalign{\smallskip}
                  & $66\%$ & MPRS & $3.72$ & $ 15.11 $ & $ 4.80 $&   $ 65.18$&    $ 0.005 $\\
                  &        & IDW   & $4.21$ & $ 17.23 $ & $ 5.27$&  $ 60.77$&    $ 0.001 $\\
\hline
\end{tabular}
\end{table}

\section{Discussion and Conclusions}
\label{sec:conclusion}
We proposed a machine learning method (MPRS) based on the modified planar rotator for spatial regression. The MPRS method is inspired from statistical physics spin models and is applicable to scattered and gridded data. Spatial correlations are captured via distance-dependent short-range spatial interactions.  The method is inherently nonlinear, as evidenced in the energy equations~\eqref{Hamiltonian_MPRS} and~\eqref{eq:Phi-ij}. The model parameters and hyper-parameters are fixed to default values for increased computational performance.  Training of the model is thus restricted to equilibrium  relaxation which is achieved by means of conditional Monte Carlo simulations. 



The MPRS prediction performance is competitive with standard methods, such as ordinary kriging and inverse distance weighting. For  data that are spatially smooth or close to the Gaussian distribution, standard prediction methods overall show better prediction performance. The relative MPRS prediction performance improves for data with rougher spatial or temporal variation, as well as for strongly non-Gaussian distributions.  For example, MPRS performance is quite favorable for daily precipitation time series which involve large number of zeros. 

The MPRS method is non-parametric: it does not assume a particular distribution, grid structure or  dimension of the data support. Moreover, it can operate fully autonomously, without user input (expertise). A significant advantage of MPRS is its superior computational efficiency and scalability with respect to sample size. These features render it suitable for massive data sets. The required CPU time increases only linearly with the size of the prediction set and does not depend on the sample size. The high computational efficiency also benefits from the full vectorization of the MPRS prediction algorithm. Thus, data sets involving millions of nodes can be processed in terms of seconds on a typical personal computer.

Possible extensions  include  generalizations of the MPRS Hamiltonian~\eqref{Hamiltonian_MPRS}. For example, spatial anisotropy can be incorporated by introducing directional dependence in the exchange interaction form~\eqref{eq:exp_dec}. Another extension could include an external polarizing field to generate spatial trends. Such a generalization would involve additional parameters that could be estimated by means of cross-validation, increasing the computational cost. This approach was applied to the MPR method on 2D regular grids, and it was shown to achieve substantial benefits in terms of improved prediction performance~\citep{vzukovivc2023spatial}. 
Finally,  the training of MPRS can be extended to include the estimation of optimal values for the model parameters. This will improve the predictive performance at the expense of some computational cost.

\backmatter

%

%
%

\section*{DECLARATIONS}

\bmhead{FUNDING}
This study was funded by the Scientific Grant Agency of Ministry of Education of Slovak Republic (Grant No. 1/0695/23) and the Slovak Research and Development Agency (Grant No. APVV-20-0150).

\bmhead{AVAILABILITY OF DATA AND MATERIALS}
The gamma dose rate, Jura, and Walker lake data sets can be downloaded from \url{https://wiki.52north.org/AI\_GEOSTATS/AI\_GEOSTATSData/}. The latent heat release data can be downloaded from \url{https://disc.gsfc.nasa.gov/datasets/TRMM_3A12_7/summary}. The used time series can be  downloaded from \url{https://pkg.yangzhuoranyang.com/tsdl/}.
The 3D  soil data  can be  downloaded from \url{http://www.leg.ufpr.br/doku.php/pessoais:paulojus:mbgbook:datasets}.

Our Matlab code is freely downloadable from \url{https://www.mathworks.com/matlabcentral/fileexchange/135757-mprs-method}. For IDW and for OK interpolation we used the Matlab codes~\citep{tovar14}  and~\citep{schwan10} respectively; both  were downloaded from the Mathworks File Exchange site.  

\bmhead{COMPETING INTERESTS}
The authors declare no conflict of interest. 
All authors certify that they have no affiliations with or involvement in any organization or entity with any financial interest or non-financial interest in the subject matter or materials discussed in this manuscript.

 \bibliography{bibliography}

\end{document}